\documentclass[journal]{vgtc}                

\usepackage{mathptmx}
\usepackage{graphicx}
\usepackage{epstopdf}
\usepackage{times}
\usepackage[lined,algonl,boxed]{algorithm2e}
\usepackage{amssymb,amsmath}
\usepackage{multirow}
\usepackage{subfigure}
\usepackage{paralist}
\usepackage{tikz}
\usepackage{multirow}
\usepackage{color}
\usepackage{ulem}
\usepackage{setspace}
\usepackage{microtype,hyphenat,balance}
\usepackage{stfloats}
\usepackage{array}
\usepackage{url}
\usepackage{bm}
\newcommand{\junz}[1]{\textcolor{black}{#1}}
\newcommand{\junzz}[1]{\textcolor{black}{#1}}

\definecolor{mypink}{RGB}{255, 100, 130}
\newcommand{\sjx}[1]{\textcolor{black}{#1}}
\newcommand{\lcx}[1]{\textcolor{black}{#1}}
\newcommand{\doc}[1]{\textcolor{black}{#1}}

\onlineid{215}

\vgtccategory{Research}

\vgtcinsertpkg

\title{Towards Better Analysis of Deep Convolutional Neural Networks}

\author{Mengchen Liu, Jiaxin Shi, Zhen Li, Chongxuan Li, Jun Zhu, Shixia Liu}

 \authorfooter{
\item
M. Liu, J. Shi, Z. Li, C. Li, J. Zhu, and S. Liu are with Tsinghua University.

 }

\abstract{
Deep convolutional neural networks (CNNs) have achieved breakthrough performance in many pattern recognition tasks such as image classification.
However, the development of high-quality deep models typically relies on a substantial amount of trial-and-error, as there is still no clear understanding of when and why a deep model works.
In this paper, we present a visual analytics approach for better understanding, diagnosing, and refining deep CNNs.
We formulate a deep CNN as a directed acyclic graph.
Based on this formulation, a hybrid visualization is developed to disclose the multiple facets of each neuron and the interactions between them.
In particular, we introduce a hierarchical rectangle packing algorithm and a matrix reordering algorithm to show the derived features of a neuron cluster.
We also propose a biclustering-based edge bundling method to reduce visual clutter caused by a large number of connections between neurons.
We evaluated our method on a set of CNNs and the results are generally favorable.
}


\teaser{
\centering
\centering
  \includegraphics[width=0.8\linewidth]{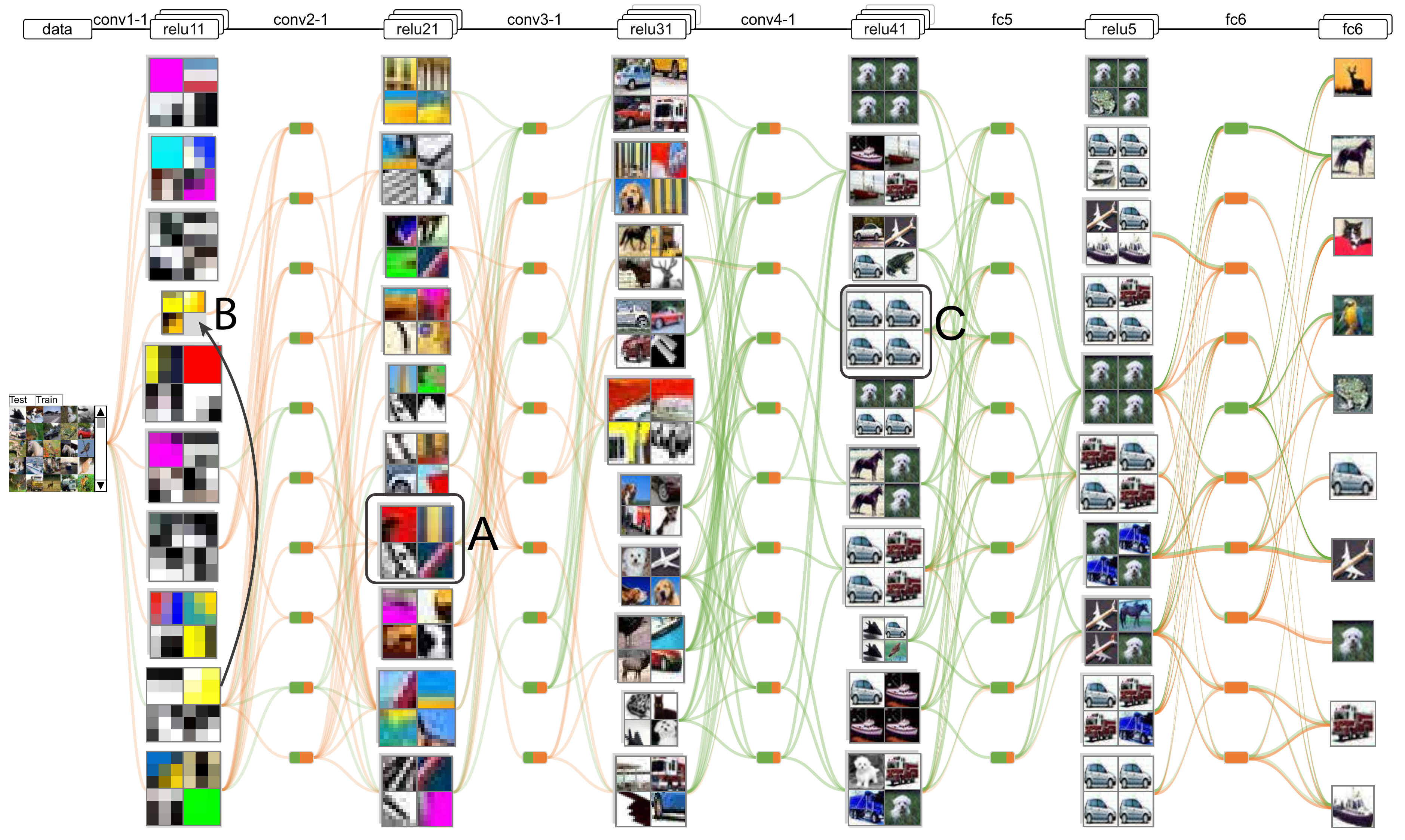}
    \vspace{-3mm}
  \caption{
    CNNVis, a visual analytics toolkit that helps experts understand, diagnose, and refine deep CNNs.
  }
\vspace{-1mm}
\label{fig:basecnn}
}

\begin{document}


{
\fontsize{8}{8} 

\firstsection{Introduction}
\maketitle
Deep convolutional neural networks (CNNs) have demonstrated significant improvements over traditional approaches in many pattern recognition tasks~\cite{Lecun2015_Survey}, such as speech recognition~\cite{Mohamed2012_Speech, Seide2011_Sppech}, image classification~\cite{He2015_CVPR, Krizhevsky2012_NIPS}, and video classification~\cite{Karpathy2014_CVPR,Zeiler2014_ECCV}.
More recently, deep CNNs have been employed as function approximators in deep reinforcement learning to extract robust representations and help make decisions, which has led to human-level performance in intelligent tasks such as Atari games~\cite{Mnih2015_Go} and the game of Go~\cite{Silver2016_Go}.
However, \junz{a deep CNN is often treated as a ``black box" model because of its incomprehensible functions and unclear working mechanism}~\cite{Bengio2013_PAMI}.
It is \junz{generally} difficult for machine learning experts to understand the role of each component (neuron, connection) due to the large number of interacting, non-linear parts in a CNN.
Without a clear understanding of how and why these networks work, the development of \junz{high-performance} models {\junz typically relies on} a substantial amount of trial-and-error~\cite{Bengio2009_Survey,Bengio2013_PAMI,Yosinski2015_Arxiv},
which is time-consuming.
For example, training a single deep CNN on a large dataset may take several days or even weeks.

There are two technical challenges to understanding and analyzing deep CNNs.
First, a CNN may consist of \junz{tens or} hundreds of layers (depth), thousands of neurons (width) in each layer, as well as millions of connections between neurons.
Such large CNNs are hard to study due to the sizes involved.
Second, CNNs consist of many functional components whose values and roles are not well understood either as individuals or as a whole~\cite{Jarrett2009_ICCV}.
In addition, how the non-linear components interact with each other and \junz{with} other linear components in a CNN is not well understood by experts.
In most cases, it is hard to summarize reusable knowledge from a failed or successful training case and transfer it to \junz{the development of} other relevant deep learning \junz{models}.\looseness=-1

To tackle these challenges, we have developed an interactive, visual analytics system called CNNVis,
which aims to help machine learning experts better understand, diagnose, and refine CNNs.
Based on the characteristics of a deep CNN,  we formulate it as a directed acyclic graph (DAG), in which each node represents a neuron and each edge represents the connection between a pair of neurons.
In order to visualize a large CNN, we first cluster the layers in the network and select a representative one from each \junz{layer} cluster.
Then we cluster neurons in each representative layer and select several representative neurons from each neuron cluster.
On the basis of the DAG representation, we develop a hybrid visualization to disclose the interactions between neurons and the multiple facets of each neuron by indicating its role for different types of images.
In particular, a hierarchical rectangle packing algorithm is developed to show the derived features of the neuron cluster.
We also design a matrix reordering algorithm based on the Held-Karp algorithm (state compression dynamic programming)~\cite{Held1962} to demonstrate the cluster patterns in the activations of each neuron cluster.
Here, the activation is the output value of a neuron, which is determined by the activation function that transforms the input value to the output value of the neuron.
Moreover, we propose a biclustering-based edge bundling method to reduce the visual clutter caused by the large number of connections between neurons.

%
%
%

In this work, we use image classification as an example and conduct three case studies with machine learning experts.
In particular, the first case study helps to illustrate the influence of the CNN model structure on performance, especially the depth and width of a CNN;
the second case study demonstrates how CNNVis helps diagnose the potential issues of a failed training case;
and the last case study illustrates how CNNVis helps refine a CNN to improve its performance.
The case studies have shown that with CNNVis, experts can better explore and understand a deep CNN, including the role of each neuron and the connections between \junz{neurons}.
For example, if a CNN suffers from overfitting in the training process, some neurons learn the same feature(s), which indicates that some of them are redundant.
Furthermore, experts can diagnose the potential issues of the model structure and refine a CNN, which enables more rapid iteration and faster convergence in model construction.

The key technical contributions of this work are:
\begin{compactitem}
\item \textbf{\normalsize A visual analytics system} that helps experts understand, diagnose, and refine deep CNNs.
\item \textbf{\normalsize A hybrid visualization} that combines a DAG with rectangle packing, matrix visualization, and a biclustering-based edge bundling method.
\end{compactitem}

\section{Related Work}\label{sec:related-work}
%
To help experts better understand a deep CNN, researchers in the field of computer vision have \junz{made} efforts to illustrate the learned features of each neuron, which is represented by part of a real image or a synthesized image.
Existing methods can be classified into two categories\junz{, namely,} code inversion~\cite{Dosovitskiy2015_Arxiv, Mahendran2015_CVPR, Zeiler2014_ECCV} and activation maximization~\cite{Erhan2009_Report, Nguyen2016_Arxiv, Simonyan2013_Arxiv, Wei2015_Arxiv, Yosinski2015_Arxiv}.

Code inversion methods synthesize an image from the activation vector of a specific layer, which is produced by a real image.
For example, Zeiler et al.~\cite{Zeiler2014_ECCV} utilized a multi-layered Deconvolutional Network~\cite{Zeiler2011_ICCV} to project the activations onto the input pixel space.
However, simple projection without considering any prior will produce images that do not resemble natural images.
To solve this problem, Mahendran et al.~\cite{Mahendran2015_CVPR} proposed incorporating several natural image priors like $\alpha$-norm and total variation to make the reconstructed images more realistic.
Recently, Dosovitskiy et al.~\cite{Dosovitskiy2015_Arxiv} trained a CNN to reconstruct the images from the activations.
They argued that a CNN can learn more powerful priors and have better performance than that of the manually defined ones.\looseness=-1 

Activation maximization \junz{methods aim} to find an image that maximally activates a given neuron.
It can be modeled as an optimization problem over the image space.
Similar to code inversion methods, natural image priors are necessary as regularization during the optimization to obtain realistic images.
As a result, most activation maximization methods focus on defining the regularization term using natural image priors~\cite{Erhan2009_Report, Nguyen2016_Arxiv, Wei2015_Arxiv, Yosinski2015_Arxiv}.
For example, Erhan et al.~\cite{Erhan2009_Report} constrained the L2-norm of the image to be constant.
Yosinski et al.~\cite{Yosinski2015_Arxiv} defined several more powerful priors, including Gaussian blur, clipping pixels with a small norm, and clipping pixels with a small contribution.

The aforementioned methods employ a grid-based representation to display the neuron features.
Although they can show the reconstructed intermediate states of each layer, they fail to disclose the inner working mechanism of CNNs, especially the role of each neuron for different types of images and the interactions between neurons.
Unlike these methods, we formulate a deep CNN as a DAG.
Based on the DAG representation, we have developed a hybrid visualization that consists of rectangle packing, matrix ordering, and biclustering-based edge bundling.
Empowered by the hybrid visualization, our visual analytics approach well discloses the multiple facets of each neuron as well as the interactions between them, which is very useful to understand the inner working mechanism of a deep CNN.
In the field of visualization, researchers have achieved a great deal of success in modeling domain-specific data as a DAG.
Typical data includes dynamic relationships between entities~\cite{Liu2013_TVCG, Tanahashi2015_TVCG, Tanahashi2012_TVCG}, temporal topic data~\cite{Cui2014_TVCG, Gad2015_TVCG, Sun2014_TVCG, Xu2013_TVCG}, temporal event sequences~\cite{Wongsuphasawat2012_TVCG}, evolving egocentric network~\cite{Wu2016_TVCG}, and the information of musicians~\cite{Janick2016_TVCG}.
Researchers have also developed a set of visualizations to reveal patterns learned from the above data.
However, none of these visualizations can be directly applied to illustrate deep CNNs because they lack a way to efficiently handle a large CNN that consists of \junz{tens or} hundreds of layers, thousands of neurons in each layer, and millions of connections between neurons.
In addition, these methods do not disclose the multiple facets of each neuron by showing its role for different types of images.\looseness=-1

Another relevant method is BiSet~\cite{Sun2016_TVCG}, which employs biclustering-based edge bundling to explore coordinated relationships between entity sets.
In BiSet, each edge is unweighted;
while in a deep CNN, each edge has a weight to indicate the impact of the input on the output.
If we simply convert a CNN to an unweighted graph and then use the biclustering method in BiSet, we may lose some important biclusters.
To solve this problem, we have developed a weighted biclustering method based on the Apriori algorithm, which is an algorithm for frequent item set mining~\cite{Agrawal1994}.

Similar to our work, Tzeng et al.~\cite{Tzeng2005} also employed a DAG to represent a neural network.
Although this visualization method can illustrate the interactions between neurons, it suffers from serious visual clutter when handling large neural networks.
To address this issue, we first cluster the layers in the network and select a representative from each \junz{layer} cluster.
Then we cluster neurons in each representative layer and select several representative neurons from each neuron cluster.
Each node in the DAG represents a neuron cluster and the edge between nodes represents the connection between the neurons in each cluster.
We have also proposed a biclustering-based edge bundling method to reduce visual clutter caused by a large number of connections between neurons.\looseness=-1

\section{Background}
\label{sec:background}

\begin{figure}[ht]
  \centering
  \includegraphics[width=\linewidth]{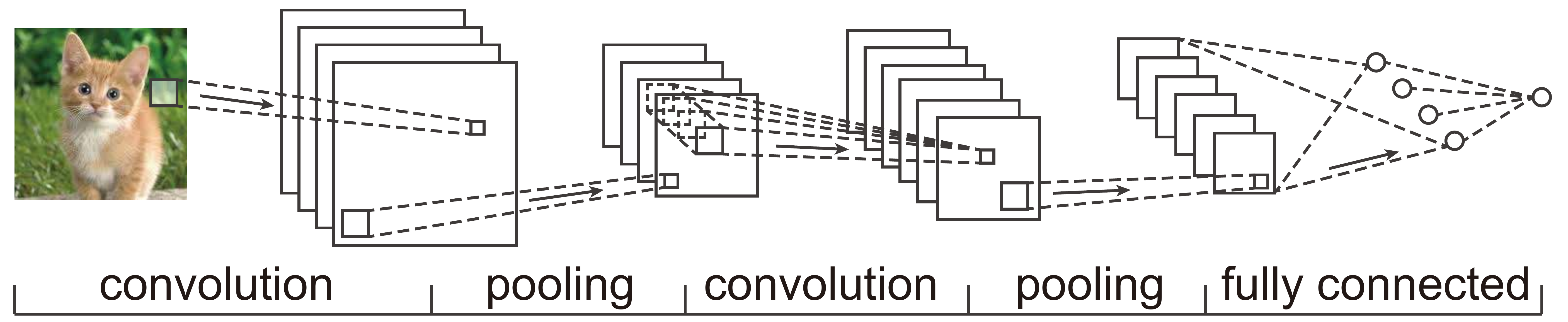}
  \vspace{-5mm}
  \caption{The typical architecture of a CNN.}\looseness=-1
  \label{fig:lenet}
  \vspace{-3mm}
\end{figure}

\begin{figure}[ht]
  \centering
  \includegraphics[width=\linewidth]{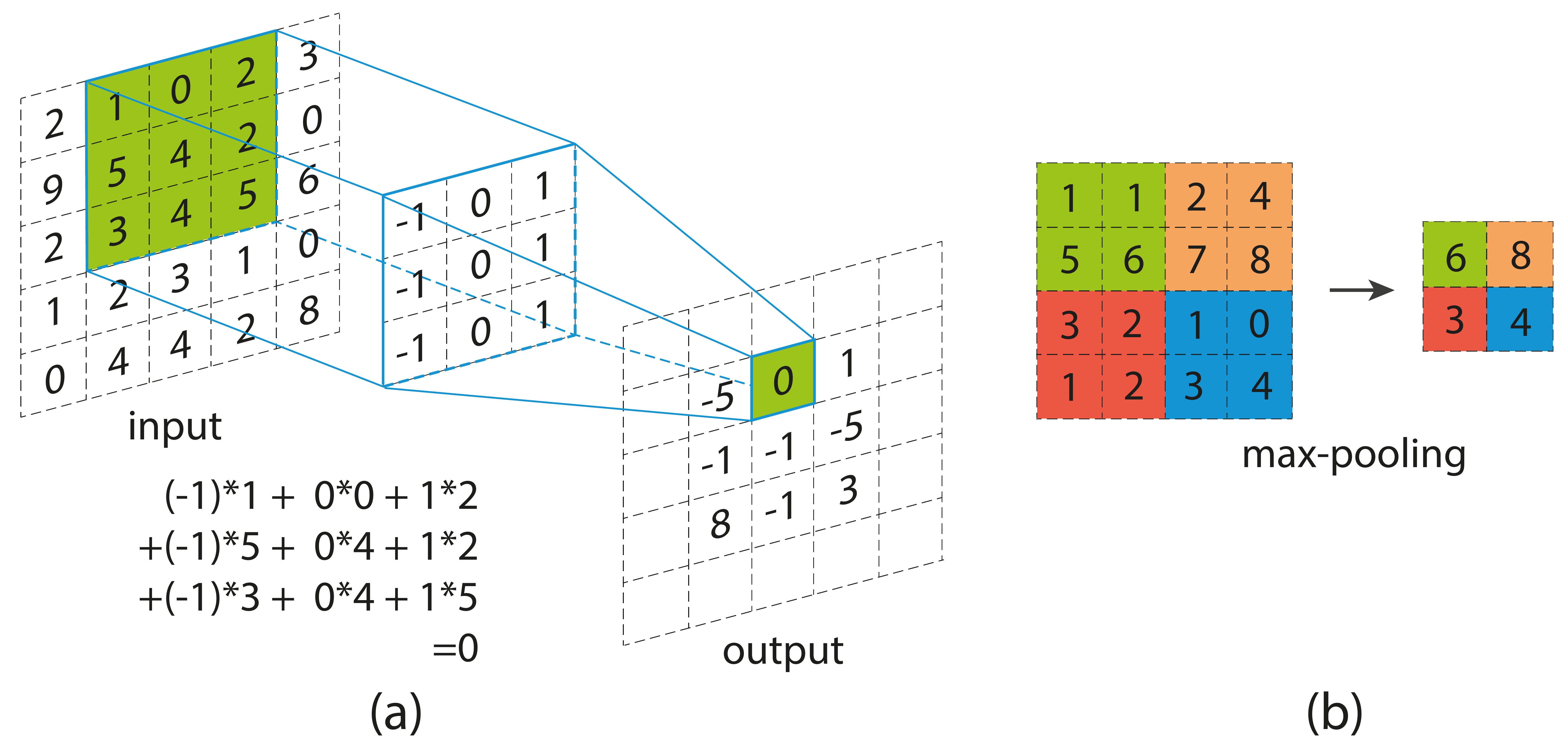}
  \vspace{-5mm}
  \caption{Illustration of convolution and max-pooling: (a) convolution; (b) max-pooling.}\looseness=-1
  \label{fig:convolution}
  \vspace{-3mm}
\end{figure}




In this section, we briefly introduce the architecture of CNNs and several basic concepts, which are useful for subsequent discussions.

CNNs are a specialized kind of neural \junz{networks} for processing data that \junz{have} a known, grid-like topology~\cite{Lecun2015_Survey}.
First, we briefly illustrate the architecture of CNNs.

\noindent \textbf{\normalsize Architecture.}
As shown in Fig.~\ref{fig:lenet}, a \junz{CNN is typically} composed of \junz{multiple} alternating convolutional and pooling layers, \junz{followed} 
by \junz{one or several} fully connected layers~\cite{Lecun2015_Survey}.
CNNs exploit local correlations by enforcing a local connectivity pattern between neurons of adjacent layers, namely, the inputs of neurons in the current layer come from a subset of neurons in the previous layer.
This hierarchical architecture allows convolutional neural networks to extract more and more abstract representations from the lower layer to the higher layer.
Fig.~\ref{fig:lenet} illustrates the architecture of a CNN that contains two convolutional and two pooling layers followed by one fully connected layer.
Next, we introduce the key components of CNNs.

\noindent\textbf{\normalsize Convolution.}
A convolution operation is performed as a window of weights slides across an image, where an output pixel produced at each position is a weighted sum of the input pixels covered by the window.
The weights that parameterize the window remain the same throughout the scanning process.
Therefore, convolutional layers can capture the shift-invariance of visual patterns and learn robust features.
The convolution operation is illustrated in \junzz{Fig.}~\ref{fig:convolution}(a), where the value of the green pixel in the output is the weighted sum of the pixels in the green region of the input.


\noindent\textbf{\normalsize Activation Function.}
\junz{An} activation function is a non-linear transformation that has been traditionally used in neural networks.
For convolutional layers, the activation function is applied after the convolution operation.
By employing activation functions, CNNs avoid learning trivial linear combinations of the inputs.
\junz{One of the most popular activation functions} is the rectified linear unit (ReLU)~\cite{nair2010rectified}.
This activation function is a piecewise linear function that prunes the negative part of the input to zero and retains the positive part:
\begin{equation}
\label{eq:3.1}
f(x) = \max(0, x).
\end{equation}

For classification tasks using probability-based loss functions like cross-entropy (see the loss function part), we often require the network output to be a vector of label probabilities, which add up to 1.
The softmax function is a special kind of activation function that satisfies this constraint:
\begin{equation}
\label{eq:3.2}
f(x)_i = \frac{e^{x_i}}{\sum_j e^{x_j}},
\end{equation}
where $x$ is the result of the linear transformation through the weights in the output layer.
After applying softmax, the output $f(x)$ is normalized to add up to 1.

\noindent\textbf{\normalsize Pooling.}
A pooling operation computes a specific norm over small regions on the input, which achieves some level of translation invariance.
This operation aggregates small pitches of pixels and thus downsamples the image features from the previous layer,
which significantly reduces the computational cost when the neural network is deep.
The most commonly used pooling operation in CNNs is max-pooling, which outputs the maximum ($L_{\infty}$ norm) pixel value of the input region (Fig.~\ref{fig:convolution}(b)).


\noindent\textbf{\normalsize Normalization.}
Normalization is an optional operation in CNNs.
It is used to speed up the convergence of the training process and reduce the probability of getting stuck in local optima~\cite{ioffe2015batch}.
This operation works by normalizing the output of certain layers through linear or non-linear operations.
Many normalization methods have been developed for CNNs such as batch normalization~\cite{ioffe2015batch}.

\noindent\textbf{\normalsize Loss Function.}
 A loss function is used to evaluate the difference between the output of a CNN and a true image label (i.e.\junz{,} the loss).
The aim of training a CNN is to minimize the value of the loss function.
It is usually achieved with stochastic gradient decent \cite{bottou-91c}, an optimization method that calculates the gradient of the loss function with respect to the weight of each edge in the network and then updates the weight according to the computed gradient.
Among various kinds of loss functions, the cross-entropy \junz{loss} along with softmax output activation\junz{s} is most commonly used for classification tasks.
This loss calculates the cross entropy between the ground truth distribution and the predicted distribution of CNNs\junz{:}
\begin{equation}
\label{eq:3.3}
l_{c}(o, t) = \sum_{i=1}^n-t_i\log o_i,
\end{equation}
where $n$ is the number of classes, $o$ \junz{denotes} the network output that is represented by a vector of probabilities for each class label, and $t$ is a one-hot vector for the true label of the current input.

Another commonly used loss function is the hinge loss\junz{, which} measures the difference between the score of the correct class and the score of the predicted class.
It is likely to have better performance on object detection tasks \cite{girshick2014rich}.
For the sake of simplicity, we introduce the hinge loss function for two classes, which is defined by:
\begin{equation}
\label{eq:3.4}
l_{h}(o, t) = max(0, 1-t \cdot o),
\end{equation}
where $t \in \{-1, +1\}$ is the class label, and $o$ is the real-valued class score produced by the network. The extension to a multi-class hinge loss can be found in~\cite{li2015max}.

\section{CNNVis}

\label{sec:requirments}

\begin{figure*}[ht]
  \centering
  \includegraphics[width=\linewidth]{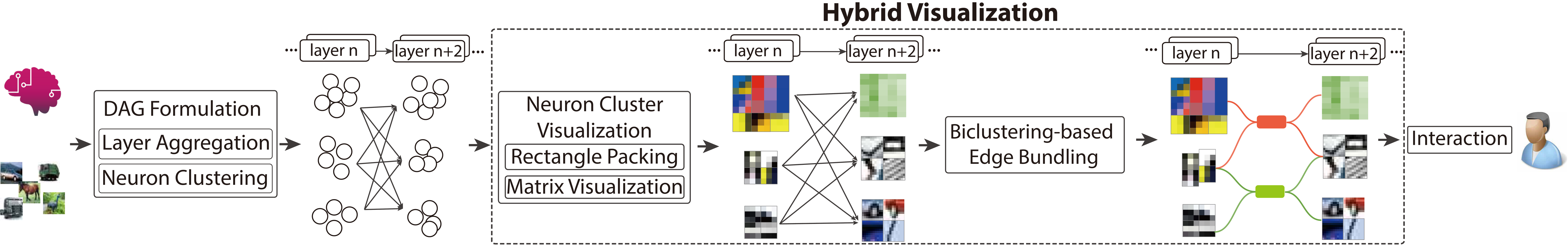}
  \vspace{-7mm}
  \caption{CNNVis pipeline.}\looseness=-1
  \label{fig:system}
  \vspace{-4mm}
\end{figure*}

CNNVis was designed with a team of deep learning experts (six researchers) over the course of twelve months.
For simplicity's sake, we denote these experts as $E_i$ ($i=1,2,\cdots,6$).
We held discussions every two weeks.
Three co-authors of this paper are also members of the team.
The development of CNNVis was triggered by their need to make sense of the inner mechanisms of deep CNNs and their dissatisfaction with the state-of-the-art toolkits.

Common deep learning frameworks include Caffe~\cite{Jia2014_Caffe}, Theano~\cite{Bergstra2010_Theano}, Torch~\cite{Collobert2002_Torch}, and TensorFlow~\cite{Abadi2016_tensorflow}.
Researchers can use these frameworks to train, debug, and deploy CNNs.
Although the deep learning frameworks output high-level statistical information, such as the training loss, as well as debugging information, such as the learned features of neurons and the gradients of weights,
it fails to disclose the role of each neuron for different categories of images and how the neurons work together.
Accordingly, if a training process fails, it is still hard for experts to figure out what is wrong with the current model design.
The experts have expressed that the development of high-quality CNN models is usually a trial-and-error procedure.
As a result, they need a toolkit that can help them better understand the inner \junzz{mechanism} of CNNs, including the role of each neuron for the different categories of images as well as the interactions between neurons.
This will allow them to summarize reusable knowledge from a failed or successful training case and transfer it to other relevant deep learning tasks.\looseness=-1

\subsection{Requirement Analysis}
\label{sec:require}
We identified the following high-level requirements based on our discussions with \junzz{the} experts and previous research.

\noindent \textbf{\normalsize R1 - Providing an overview of the learned features of neurons.}
All the experts commented that an overview of the learned features of neurons is necessary to begin their analysis (e.g., diagnosis or refinement of the model).
They usually examine the quality of each learned feature layer by layer to discover potential problems.
However, such an examination can be very difficult for a deep CNN with \junz{tens or} hundreds of layers and thousands of neurons in a layer.
As a result, they stated the need to cluster neurons into clusters so they can gain a quick overview of the learned features of each cluster.

\noindent \textbf{\normalsize R2 - Interactively modifying the neuron clustering results.}
Since the clustering algorithm may be imperfect and different users may have different needs, experts need to interactively modify the clustering results based on their knowledge.
Expert $E_2$ commented that when examining the training \junz{results} of a CNN, he found a neuron for detecting a color patch in a cluster that mainly consists of neurons for detecting stripes with various orientations.
To increase the clustering accuracy and better compare these clusters, he moved the neuron to a cluster that mainly consists of neurons for detecting color patches.


\noindent \textbf{\normalsize R3 - Exploring multiple facets of neurons.}
Previous work mainly \junz{focused} on visualizing the learned features of neurons.
In addition to this feature, the experts also requested viewing other facets of neurons.
For example, expert $E_1$ said, ``In addition to the learned features, other numerical features such as activation (of a neuron) can also help me better understand its role in a classification task.''
During the discussion, we gradually identified that the major facets of interest are the learned features (all the experts), activations ($E_1$, $E_3$, $E_4$, $E_5$, $E_6$), and contributions to the final result (all the experts).
Visually illustrating them can help experts gain a more comprehensive understanding of the roles of neurons.\looseness=-1

\noindent \textbf{\normalsize R4 - Revealing how low-level features are aggregated into high-level features.}
In a CNN, neurons in lower layers learn to detect simple features such as stripes or corners, neurons in middle layers learn to detect a part of an object, and neurons in higher layers learn to detect a concept (e.g., a cat).
This is achieved with a local connectivity pattern between neurons of adjacent layers, which means the inputs of neurons in layer \emph{m} are from a subset of neurons in layer m-1.
As a result, the experts wanted to learn how neurons in adjacent layers interact with each other and aggregate the low-level features into high-level features.
Previous research has also shown that analyzing such connections can help experts understand how a large number of non-linear parts interact with each other~\cite{Tzeng2005}.
A large CNN may contain millions of connections between neurons.
If we display all them, it is difficult to discern individual connection due to visual clutter caused by excessive edges and edge crossings.
Thus, the experts required to examine the major trends among these connections.



\noindent \textbf{\normalsize R5 - Examining the debugging information.}
In the discussions, the experts expressed the need to examine the debugging information of the deep model.
Expert $E_3$ said, ``I often examine the debugging information such as the gradients, to diagnose a training process \junz{that} failed to converge.''
In addition to gradients, showing other derived values such as the relative change of weights, has also been requested by the experts.
The debugging information is usually huge\junz{. For} example, there are millions of gradients\junz{. It} is very hard to examine them one by one and develop a full understanding.
As a result, the experts also requested having an overview of such debugging information.
This need is consistent with the findings of previous research~\cite{Bengio2013_PAMI, Glorot2010_AISTATS}.

\subsection{System Overview}

The list of requirements \junzz{have} motivated us to develop a visual analytics system, CNNVis.
It consists of the following components:
\begin{compactitem}
\item A DAG formulation module to convert a CNN to a DAG and to aggregate neurons and layers for an overview (\textbf{R1},\textbf{R4});
\item A neuron cluster visualization module to disclose the multiple facets of each neuron (\textbf{R3});
\item A biclustering-based edge bundling to reduce visual clutter caused by a large number of connections (\textbf{R4});
\item An interaction module that provides a set of interactions such as interactive clustering result modification (\textbf{R2}) and showing debug information on demand (\textbf{R5}).
\end{compactitem}

The primary goal of CNNVis is to help experts better understand, diagnose, and refine CNNs.
Fig.~\ref{fig:system} illustrates the major components needed to achieve this goal.
CNNVis takes a trained CNN and the corresponding training data set as the input.
The input CNN is formulated as a DAG with each node representing a neuron and each edge representing the connection between neurons.
To effectively present a large CNN, the DAG formulation module clusters the neurons in each layer.
The clustered DAG is then passed to the neuron cluster visualization module.
This module employs a rectangle packing algorithm to show the learned features of each neuron in a cluster and a matrix visualization to depict the activations of neurons.
After that, a biclustering-based edge bundling clusters the edges to reduce visual clutter.
Users can also interact with the generated visualization for further analysis.
For example, users can interactively modify the neuron clustering results or show the average gradient of a selected layer.


\section{DAG Formulation}
\label{sec:formulation}

\begin{figure}[ht]
  \centering
  \includegraphics[width=\linewidth]{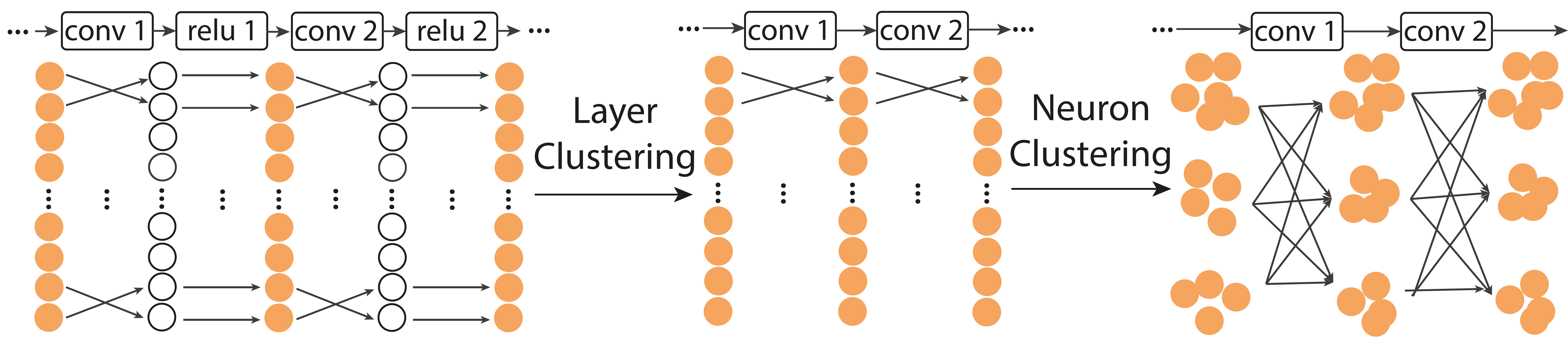}
  \vspace{-5mm}
  \caption{Illustration of the DAG formulation.}\looseness=-1
  \label{fig:dag}
  \vspace{-4mm}
\end{figure}

A CNN can be formulated as a DAG, where each node represents a neuron and each edge represents the connection between neurons.
To effectively present a large CNN with \junz{tens or} hundreds of layers and thousands of neurons in each layer, we first aggregate adjacent layers into groups.
There are several ways to do the aggregation.
For example, we can classify layers by merging two adjacent convolutional layers that have a small difference between their activation variance.
We can also divide layers into groups at each pooling layer.
In our current implementation, we employ the second one.
In addition, the experts are interested in the output of an activation layer instead of that of a convolutional layer.
As the outputs of these two layers have a one-to-one mapping relationship, we then merge these two layers and simply show the output of the activation layer (Fig.~\ref{fig:dag}).

Then we cluster the neurons in each layer,
which aims to group neurons with similar roles together.
We assume that neurons with similar activations have similar roles.
Directly using these activations to cluster the neurons is very time-consuming as there can be millions of images in the training set.
Thus, we aggregate the activations into an average activation vector over the set of classes in the training set.

In particular, suppose the training samples can be categorized into $m$ classes: $c_1, c_2,...,c_m$.
The training samples of class $c_i$ is represented by: $S_i=\{s_1^{(i)}, s_2^{(i)}, \cdots ,s_{N_i}^{(i)}\}$, where \junz{$N_i$} is the number of training samples in class $c_i$.
We first process each training sample $s_j^{(i)}$ through the network and obtain the activation of neuron $n$: $a_n(s_j^{(i)})$.
Then we calculate the average activation $a_n(c_i)$ of neuron $n$  on class $c_i$ by:
\junz{\begin{equation}
\frac{1}{N_i}\sum\limits_{j=1}^{N_i}{a_n(s_j^{(i)})}.
\end{equation}}

Next, we combine each average activation into an activation vector $\vec{a_n} = [a_n(c_1),a_n(c_2), ...,a_n(c_m)]$, which is a $m$ dimension real-valued vector.

Finally, we cluster the neurons based on the derived activation vectors.
In CNNVis, we employ two widely used clustering methods, K-Means~\cite{Marsland2015_KMeans} (parametric clustering) and MeanShift~\cite{Comaniciu2002_Clustering} (nonparametric clustering).
The second method does not require prior knowledge of the cluster number.
Thus, it is applicable to the case where experts do not know the cluster number of neurons.
To better present each neuron cluster, we select several representative neurons that are closer to the cluster centroid.
\section{Visualization}\label{sec:vis}



\subsection{Overview}

Based on the DAG formulation, we have designed a hybrid visualization (Fig.~\ref{fig:visoverview}) that visually illustrates neuron clusters (nodes) \doc{and} the connections between neurons (edges).

Each neuron cluster is represented by a \doc{large} rectangle (Fig.~\ref{fig:visoverview}A)\doc{, which} can be analyzed from multiple facets\doc{,} such as \doc{the} learned features, activations, and contributions to the final result (\textbf{R3}).
\doc{Specifically}, we \doc{have adopted} a rectangle packing algorithm to place the learned features of neurons in a neuron cluster, where each learned feature is encoded by a smaller rectangle (Fig.~\ref{fig:visoverview}B1).
\doc{Neuron activations} are visualized as a matrix visualization (Fig.~\ref{fig:visoverview}B2).
Users can switch between the rectangle packing representation and the matrix visualization to explore different facets of the neurons.


To reduce visual clutter caused by dense edges and their crossings, we have developed a biclustering-based edge bundling algorithm (\textbf{R4}).
For each layer, we first generate the biclusters between the input neuron clusters and output neuron clusters.
Inspired by BiSet~\cite{Sun2016_TVCG}, we \doc{have} also \doc{added} an ``in-between'' layer between the input neuron clusters and output neuron clusters (Fig.~\ref{fig:visoverview}C).
In this layer, each bicluster is treated as a node in the DAG and is represented by a small rectangle.\looseness=-1

In CNNVis, we employ the layout algorithm in TextFlow~\cite{Cui2011_TVCG} to calculate the position of each node (e.g., neuron cluster or a bicluster) (\textbf{R1}).
We also provide a set of interactions to facilitate deep analysis of a deep CNN (\textbf{R2}, \textbf{R5}).

Next, we \doc{will} introduce the neuron cluster visualization and biclustering-based edge bundling in details.

\begin{figure}[t]
  \centering
  \includegraphics[width=\linewidth]{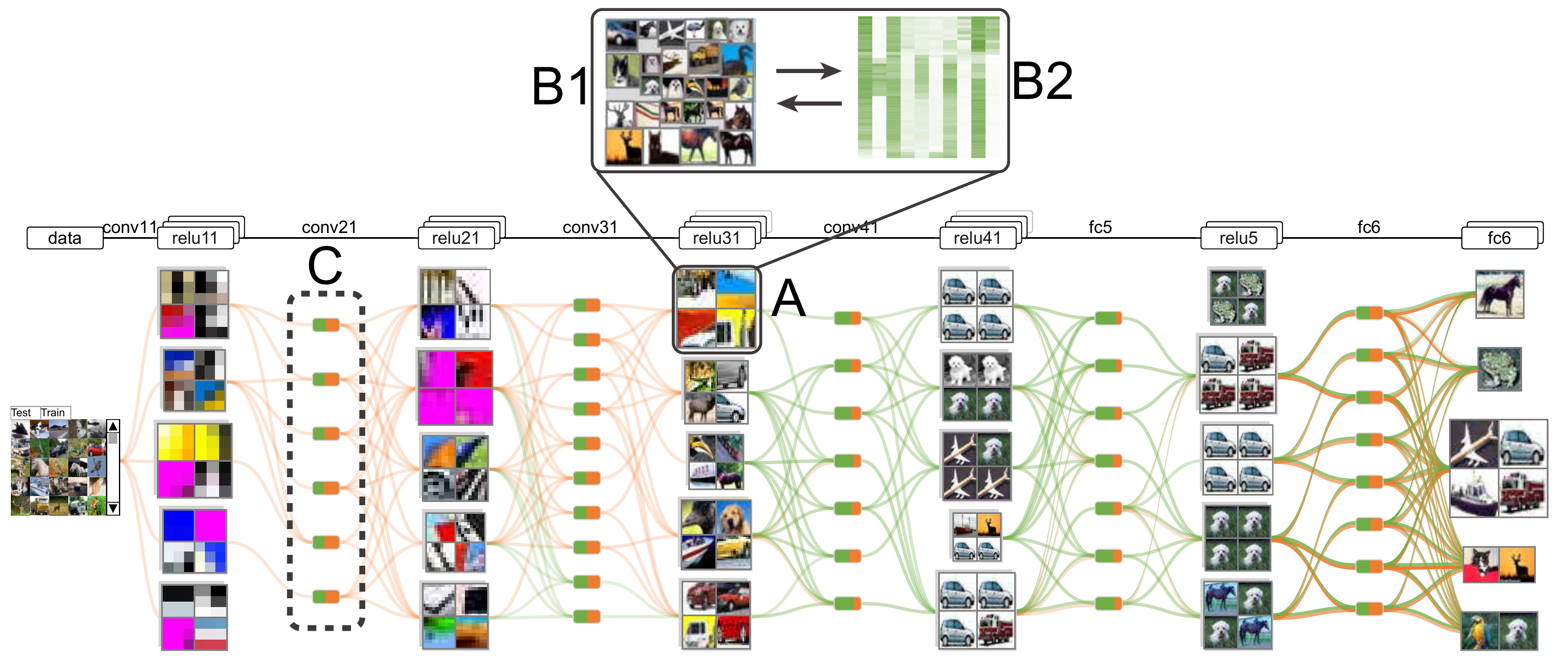}
  \vspace{-5mm}
  \caption{Visualization overview.}\looseness=-1
  \label{fig:visoverview}
  \vspace{-3mm}
\end{figure}



\subsection{Neuron Cluster Visualization}

\subsubsection{Learned Features as Rectangle Packing}


\noindent\textbf{\normalsize Computing learned features of neurons.}
We employ the method used in~\cite{girshick2014rich} to compute the learned feature of a neuron because it is fast and the results are easier to understand.
We also compute the activations of each neuron on a large set of image patches (e.g., sampled from the training set) and sort the patches in decreasing order according to their activations.
To help experts better understand the role of each neuron, we select the top-5 patches with \doc{the} highest activation scores to represent the learned feature of that neuron.
By default, we show the \doc{top} patch for a neuron and allow users to switch among these five patches.
Other methods for computing the learned feature~\cite{Mahendran2015_CVPR, Zeiler2014_ECCV} can \doc{easily} also be integrated into CNNVis.

\begin{figure}[ht]
  \centering
  \includegraphics[width=\linewidth]{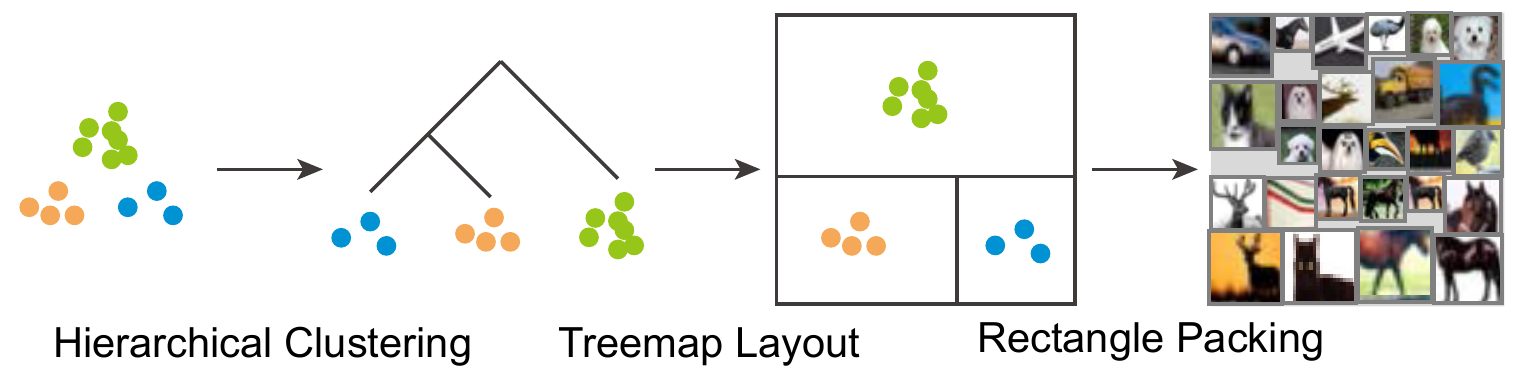}
  \vspace{-5mm}
  \caption{Illustration of hierarchical rectangle packing.}\looseness=-1
  \label{fig:packing}
  \vspace{-2mm}
\end{figure}

\noindent\textbf{\normalsize Layout.}
A straightforward way to visualize the learned features (image patches) is to employ a grid-based layout where each image patch is represented by a rectangle of the same size~\cite{Yosinski2015_Arxiv, Zeiler2014_ECCV}.
However, this method fails to emphasize the important neurons.\looseness=-1

To tackle this issue, we formulate the layout of image patches as a rectangle packing problem, \doc{aiming} to pack the given rectangles into an enclosing rectangle of a minimum area.
We use the size of an image patch to encode the importance of \junz{the} corresponding neuron because size is among the most effective visual channels~\cite{Munzner2014_Vis}.
In CNNVis, we provide several options to define the importance of a neuron, including its average or maximal activation on a set of classes and its contribution to the final result~\cite{Li2006_Margin}.

Existing rectangle packing algorithms~\cite{Huang2012_Packing, Korf2010_Packing} can handle a small number of rectangles well (e.g., 15 rectangles in less than 0.1s~\cite{Huang2012_Packing}).
However, the computing time grows exponentially as the number of packed rectangles increases (e.g., 25 rectangles in more than one hour~\cite{Huang2012_Packing}).
Since a neuron cluster may consist of hundreds or even thousands of neurons, existing rectangle packing algorithms cannot directly be applied to our visualization.

To solve this problem, we \doc{have} developed a hierarchical rectangle packing algorithm.
The basic idea of our algorithm is to divide the problem into a number of smaller sub-problems.
Each sub-problem can be efficiently solved by the state-of-the-art rectangle packing algorithm~\cite{Huang2012_Packing}.
Specifically, our algorithm contains the following steps (Fig.~\ref{fig:packing}).

\noindent\emph{\normalsize Step 1: Hierarchical clustering.}
In this step, we perform \doc{a} hierarchical clustering to divide the problem into several sub-problems that can be efficiently solved by an algorithm developed by Huang and Korf~\cite{Huang2012_Packing}.
Specifically, we start with the cluster containing all of the neurons.
Then we repeatedly split a cluster until the number of neurons in it is smaller than a threshold.
This cluster splitting is done \doc{with} a widely used graph clustering method~\cite{Newman2004_Graph}.

\noindent\emph{\normalsize Step 2: Computing the layout area for each cluster.}
Based on the hierarchical clustering result, we compute the layout area for each sub-cluster \doc{using} a Treemap layout algorithm~\cite{Johnson1991_VIS}.

\noindent\emph{\normalsize Step 3: Rectangle packing of each cluster.}
In this step, we compute the position and size for each image patch \doc{using} the state-of-the-art rectangle packing algorithm~\cite{Huang2012_Packing}.

\subsubsection{Activations as Matrix Visualization}

In our first prototype, we simply encode the activation of a neuron \doc{according to} its size.
However, the experts were not satisfied with that design because it failed to help them compare the roles of the neurons for different classes of images.
To allow experts to compare different neurons, we stack the average activation vectors of neurons into an activation matrix, where each row is an average activation vector of a neuron.
Accordingly, a matrix visualization is employed to visually illustrate the activation of \doc{the} neurons.
In particular, the color of a cell in the $i$-th row and $j$-th column represents the average activation of the $i$-th neuron $n_i$ \junz{in} class $c_j$.

This design was then presented \doc{to experts} for evaluation.
Overall, they liked the matrix visualization that provides a global overview of the activations among different classes.
Their major concern was that the current visualization cannot reveal the cluster patterns in the activations of a neuron cluster.
To solve this problem, we developed a matrix reordering algorithm that can visually reveal cluster patterns within the data.\looseness=-1


\begin{figure}[ht]
  \centering
  \includegraphics[width=\linewidth]{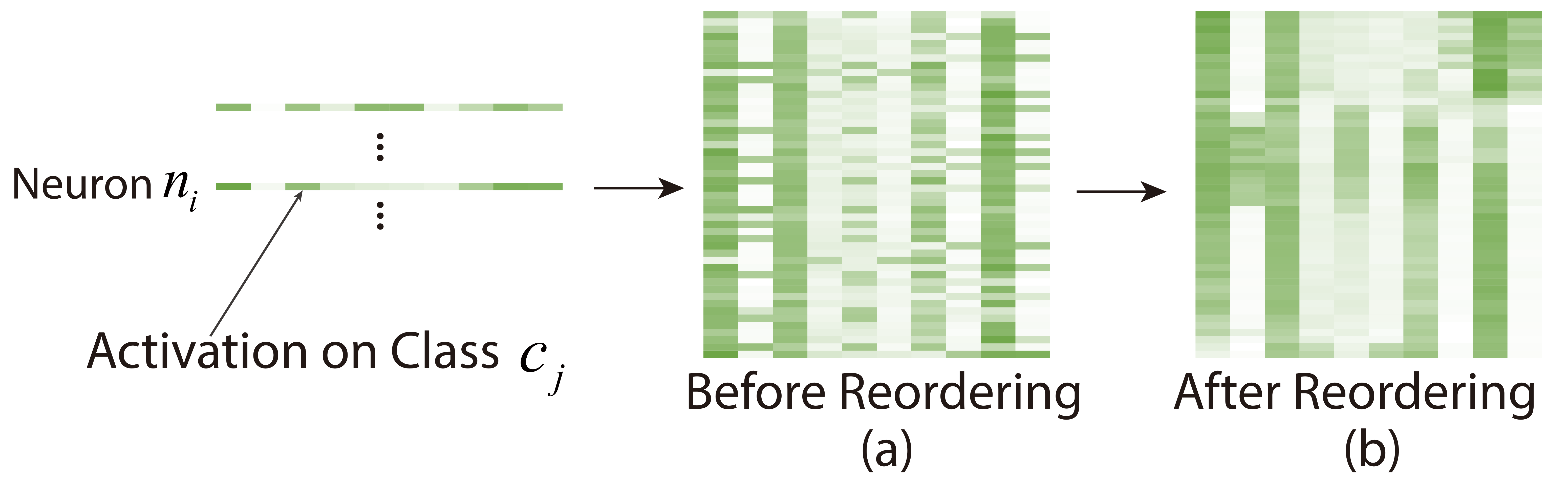}
  \vspace{-7mm}
  \caption{Illustration of matrix reordering: (a) before reordering; (b) after reordering.}\looseness=-1
  \label{fig:matrix}
  \vspace{-3mm}
\end{figure}

\noindent\textbf{\normalsize Matrix Reordering.}
The order of columns (classes) should be consistent in different neuron clusters.
Otherwise, \doc{experts are unable to} directly compare the roles of neurons in two neuron clusters because of \doc{the} different order of classes (columns).
As a result, we only reorder the rows (neurons) in the matrix.

The basic idea of our algorithm is to maximize the sum of the similarities between adjacent neurons in the matrix.
It aims to place neurons with similar activations close \doc{to} each other, and thus can reveal the cluster pattern in the neuron cluster.
Given neuron cluster $C = \{n_1, n_2,\cdots, n_{N_C}\}$,
the goal of the reordering is to find a row index $\pi(i)$ for each neuron $n_i$, to better reveal the cluster pattern in a neuron cluster.
\doc{For row} $r$ in the matrix, we denote its corresponding neuron as $n_{\pi^{-1}(r)}$.
To achieve this goal, we try to maximize the sum of the similarities between adjacent neurons in the matrix:
\begin{equation}
\label{eq:6.1}
max \sum\limits_{r=1}^{N_C-1}{sim(n_{\pi^{-1}(r)}, n_{\pi^{-1}(r+1)})},
\end{equation}
where $sim()$ is the similarity function between two neurons.
In CNNVis, we adopt the widely used cosine similarity.

This combinational optimization problem can be solved by the Held-Karp algorithm~\cite{Held1962} with a time complexity of $O(2^{N_C}\cdot N_C^2)$, where $N_C$ is the number of neurons.
The problem of directly applying it in our system is that we may have hundreds of neurons in a \junz{neuron} cluster and the running time of the algorithm is very long.
Thus, we developed a divide-and-conquer method to accelerate the algorithm, which consists of the following steps.

\noindent\emph{\normalsize Divide.} If the number of neurons in a cluster is too large to be efficiently solved \doc{via} directly running the Held-Karp algorithm, the cluster is divided into several sub-clusters by a widely used graph clustering method developed by Newman~\cite{Newman2004_Graph}.

\noindent\emph{\normalsize Conquer.} Computing the ordering of sub-clusters by running the Held-Karp algorithm.

\noindent\emph{\normalsize Combine.} Merging the ordering of sub-clusters into a global ordering.

Fig.~\ref{fig:matrix} shows one result generated \doc{using} our reordering method.
With our method, several clusters can \doc{easily be} detected.

\subsubsection{Interaction}
To better \doc{facilitate understanding of} the multiple facets of each neuron cluster, CNNVis provides a set of user interactions.

\noindent\textbf{\normalsize Interactive Clustering Result Modification.}
Since the clustering algorithm is less than perfect and experts may have different needs, we allow experts to interactively modify the clustering \doc{results} based on their knowledge (\textbf{R2}).
Inspired by NodeTrix~\cite{Henry2007_Graph}, we allow experts to drag a neuron out of a neuron cluster or to \junz{another} neuron cluster.

\noindent\textbf{\normalsize Selecting A Part of Neurons to View.}
There are thousands of neurons in a CNN.
Thus, it is necessary to allow experts to select \doc{some of the} neurons to view.
We allow users to select a set of classes and show the neurons that are strongly activated by the images in these classes.
Other irrelevant neurons are deemphasized by setting them to be translucent.

\noindent\textbf{\normalsize Switching between Facets.}
Exploring the multiple facets of neurons can help experts better understand the roles of neurons.
Thus, we allow users to switch between these facets (\textbf{R3}).
For example, users can switch to view the learned features or the activation matrix.

\subsection{Biclustering-based Edge Bundling}

Initially, we visualized each edge as a curve.
The major concern of the experts is visual clutter caused by millions of edges between nodes.

In order to reduce visual clutter, we tried two geometry-based edge bundling methods~\cite{Cui2008_Edge, Holten2009_Edge} to cluster the edges between two layers.
After interacting with CNNVis, the experts commented that this bundling method \doc{reduces} visual clutter to some extent.
However, the clusters revealed by the geometry-based bundling methods \doc{did} not help their analysis because the edges with similar weights \doc{were not clustered} together.
The experts are more interested in edges with larger absolute weights, \doc{because this} indicates that \doc{the} corresponding inputs have \doc{a} larger impact on the output.



To fulfill this requirement, we \doc{developed} a biclustering-based edge bundling method to bundle edges with \doc{both} similar and large absolute weights.
For a given layer, a bicluster is a subset of input neuron clusters and a subset of output neuron clusters.
 \doc{This method} can logically aggregate multiple individual connections and thus \doc{provides} an opportunity to visually bundle edges between neuron clusters.
Our algorithm contains the following steps (Fig.~\ref{fig:bicluster}).

\noindent\emph{\normalsize Step 1: Aggregating Connections between Neurons.}
We first calculate the strength $w_{ij}$ of the connection $e_{ij}$ between two neuron clusters, $C_i$ and $C_j$.
We denote $E = \{e_{ij}\}$ as the edge set.
An intuitive \doc{approach} is to use the average of all the weights of the edges connecting a neuron $n_s \in C_i$ and a neuron $n_t \in C_j$.
The problem \doc{with} this method is that it aggregates positive edges (edges with positive weights) and negative edges (edges with negative weights) and may result in an aggregated edge with a small weight.
This may lead to \doc{a} misunderstanding.
Thus, we calculate the strength of the connection between two neuron clusters as a two-dimensional vector $\vec{w}_{ij}=[w_{ij}^{pos}, w_{ij}^{neg}]$, where $w_{ij}^{pos}$ is the average of positive \doc{edge} weights and $w_{ij}^{neg}$ is the average of \doc{the} negative \doc{edge} weights.\looseness=-1

\noindent\emph{\normalsize Step 2: Biclustering.}
Based on the aggregation \doc{results}, we then detect biclusters between the input neuron clusters and the output neuron clusters.
 \doc{Because} experts are interested in both larger positive edges and smaller negative edges, we cannot simply convert it to an unweighted graph and perform biclustering.
Thus, we first seek the maximum value $w_{max}$ in $W = \{w_{ij}^{pos}\} \cup \{|w_{ij}^{neg}| \}$.
If $w_{max} \in \{w_{ij}^{pos}\}$, then we select the edges satisfying: $|w_{ij}^{pos} - w_{max}|<\tau$, where $\tau$ is a user defined parameter denoting the tolerance of similarity.
If $w_{max} \in \{|w_{ij}^{neg}|\}$, we then perform the similar extraction.
For these edges, we then mine the closed item sets as biclusters, where each input neuron cluster is connected to each output neuron cluster.
To mine the closed item sets, we adopt the widely used Apriori, an algorithm for frequent item set mining~\cite{Agrawal1994}.
After that, we remove the edges in the extracted biclusters from $E$ and then repeat the process until $w_{max}$ is under a user defined threshold.

\noindent\emph{\normalsize Step 3: Edge Bundling.}
In this step, we bundled the edges in the same bicluster to reduce the visual clutter.
Inspired by BiSet~\cite{Sun2016_TVCG}, we also add \junz{an} ``in-between'' layer between the input neuron clusters and \doc{the} output neuron clusters (Fig.~\ref{fig:bicluster} (c)).
In this layer, each bicluster is visualized as a rectangle.
In a bicluster, we \doc{use} two colored regions (green and red) to indicate the proportion between the number of positive edges and \doc{of} negative edges.
An edge between two neuron clusters consists of two aggregated curves (Fig.~\ref{fig:bicluster}A, and Fig.~\ref{fig:bicluster}B),
where green and red visually encode positive and negative weights, respectively.
Since experts are less interested in \doc{analyzing edges} with smaller absolute weights, they are not displayed by default.
These edges can be shown per users' request.

\noindent\textbf{\normalsize Interaction.}
The debugging information can help experts diagnose a failed training process.
In CNNVis, we allow experts to analyze the debugging information at different \doc{granularities} (\textbf{R5}).
For example, they can change the color encoding of edges to analyze the gradient of each weight.
Experts also \doc{have the option} to view the average gradient at each layer as a line chart to get an overview of the \doc{debugging} information.

\begin{figure}[ht]
  \centering
  \includegraphics[width=\linewidth]{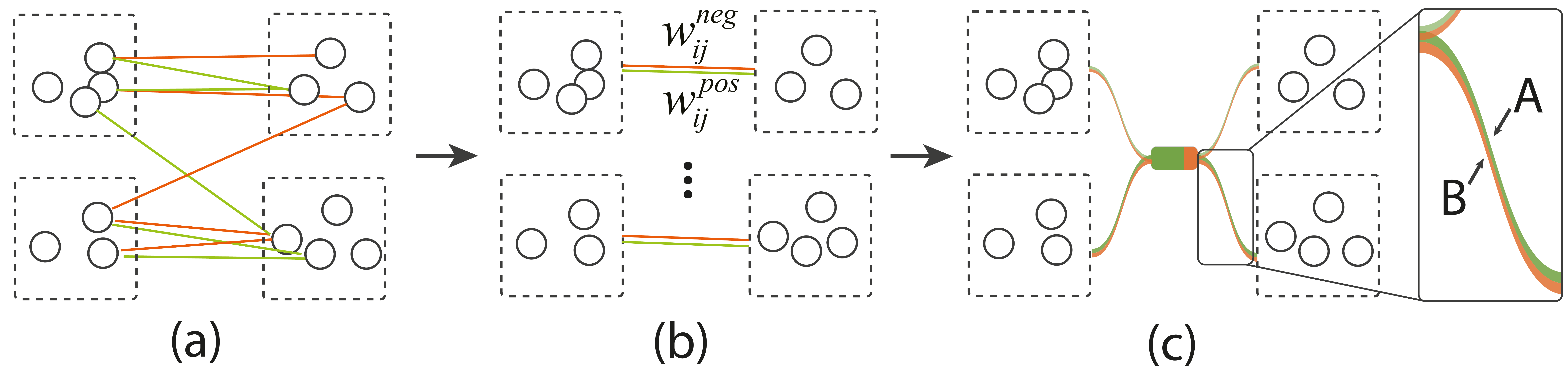}
  \vspace{-6mm}
  \caption{Illustration of biclustering-based edge bundling.}\looseness=-1
  \label{fig:bicluster}
  \vspace{-3mm}
\end{figure}

\section{Application}\label{sec:application}

In this section, we present the case studies to demonstrate how CNNVis help experts understand, diagnose and refine a CNN.

\subsection{Overview}
\label{sec:app-overview}
We have worked closely with the team \doc{of experts} to select the base CNN model and to design the case studies.

\noindent \textbf{\normalsize Base CNN.}
The base CNN \doc{was} contributed by $E_3$ of the expert team.
For \doc{brevity's sake}, we refer to the base CNN as BaseCNN.
BaseCNN \doc{was} designed based on a widely used deep CNN introduced in~\cite{Simonyan2014_Arxiv}, which is often used in image classification.
Recently, the expert team that we collaborate with has been redesigning this CNN and testing the performance of the variants.
BaseCNN consists of 10 convolutional layers and two fully connected layers.
The convolutional layers \junz{are} organized into four groups, \doc{containing} 2, 2, 3, and 3 convolutional layers, respectively.
Each group is ended with a max-pooling layer.
When designing BaseCNN, the expert employed the commonly used activation function, ReLU, and the commonly used loss function, cross-entropy.
The architecture of BaseCNN is depicted in Fig.~\ref{fig:basecnn_architecture}.

BaseCNN \doc{was} trained and tested on a benchmark image dataset, CIFAR10~\cite{Krizhevsky2009_Report},
which consists of 60,000 labeled color images of size 32$\times$32 in 10 different classes (e.g., airplane, bird, and truck), with 6,000 images per class.
The dataset is split into a training set containing 50,000 images and a test set containing 10,000 images.
Training and testing of BaseCNN \junz{are} performed \junz{under} a widely used deep learning framework, Caffe~\cite{Jia2014_Caffe}.
The BaseCNN model achieves 11.32\% error on the test set.

\noindent \textbf{\normalsize Design of Case Studies.}
We have worked closely with the expert team to design three case studies from their current research on CNNs.

First, based on BaseCNN, the expert team constructed several variants and aimed to study the influence of the network architecture \doc{on} the performance.
The \junz{experts} said that such \junz{an} analysis would help \doc{to} better understand the reason why CNNs with different architectures have different performance (Section~\ref{sec:case1}).

Second, the expert team required to diagnose a training process that \junz{failed} to converge.
For example, in one training \junzz{trial}, $E_3$ changed the output activation function and the loss function of BaseCNN.
However, the training failed.
The expert team wanted to diagnose the training process and \junzz{find} \doc{potential} issues.
This scenario triggered the second case study (Section~\ref{sec:case2}).

Finally, the expert team wanted to further improve the performance of the BaseCNN model.
To this end, the expert team \doc{decided} to examine the output of each layer from \junz{a} global overview to local details and detected \doc{a} potential direction to improve the model.
This requirement is addressed in the third case study.

Due to the page limit, we focus our report on the first two case studies.
Interested readers may refer to the attached video for the study on model refinement (third case study).

\begin{figure}[ht]
  \centering
  \includegraphics[width=\linewidth]{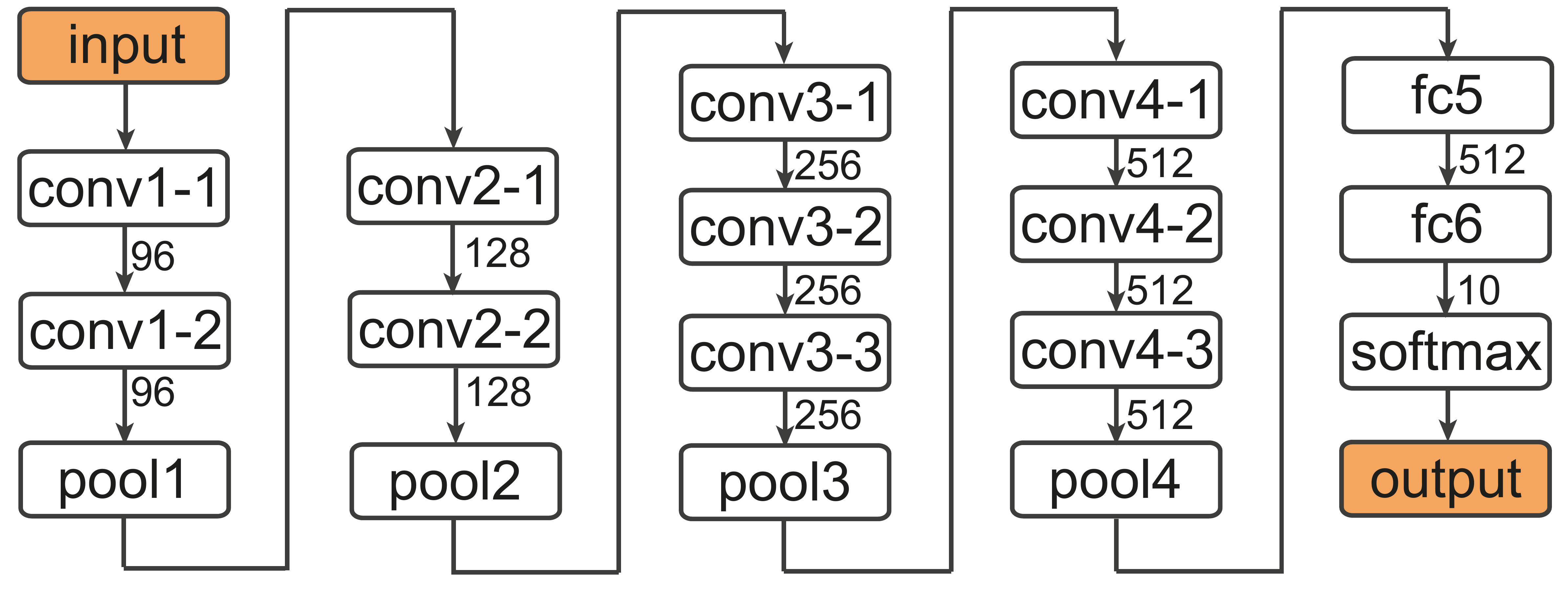}
  \vspace{-7mm}
  \caption{The architecture of BaseCNN. It contains four groups of convolutional layers and two fully connected layers.
  The number below a layer is the number of neurons in that layer.}\looseness=-1
  \label{fig:basecnn_architecture}
  \vspace{-3mm}
\end{figure}

\begin{figure}[ht]
  \centering
  \includegraphics[width=\linewidth]{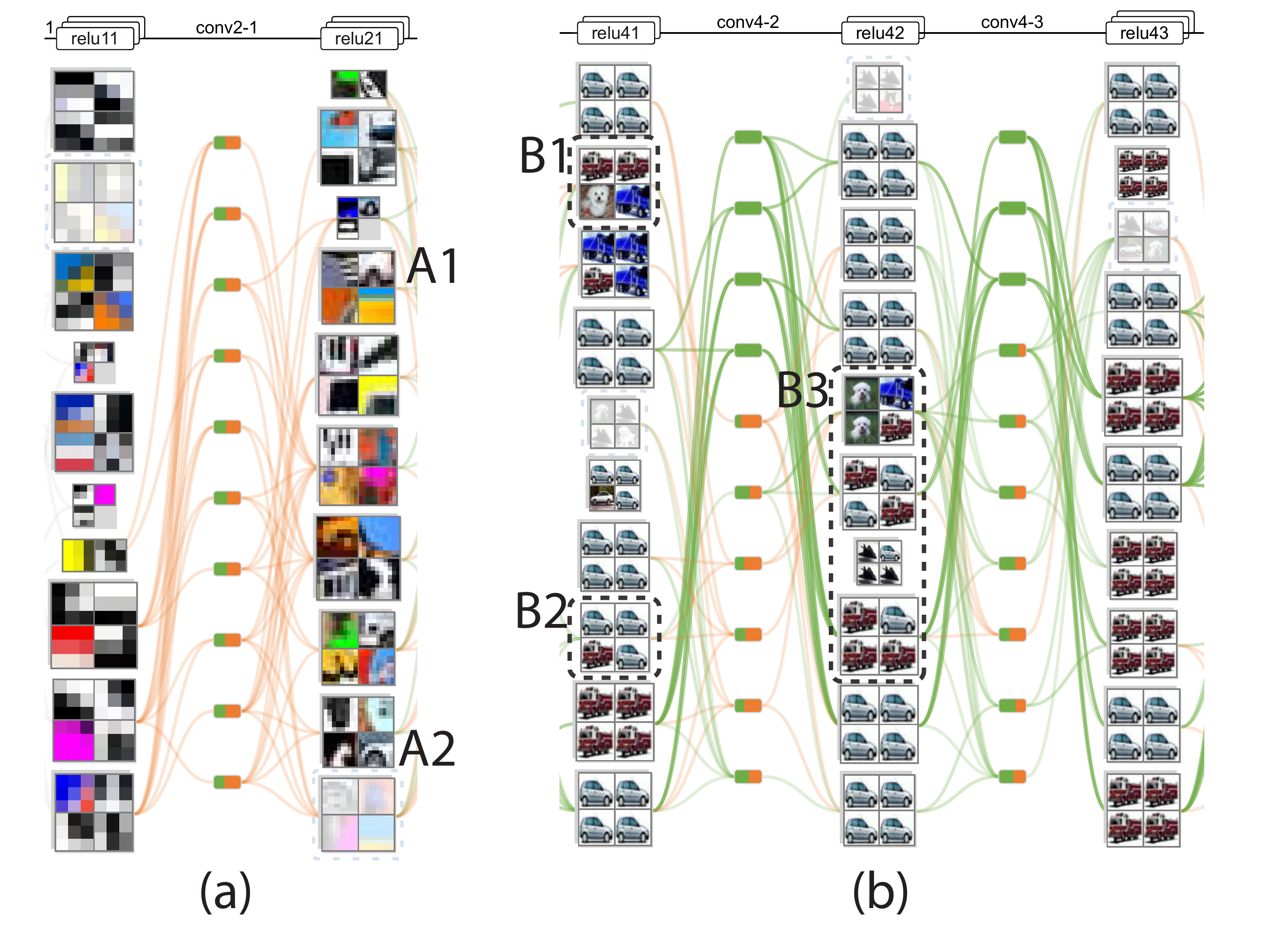}
  \vspace{-6mm}
  \caption{Learned features of BaseCNN: (a) low level feature; (b) high level features.}\looseness=-1
  \label{fig:depth}
  \vspace{-3mm}
\end{figure}

\subsection{Case Study: Influence of Network Architecture}
\label{sec:case1}

This case study was a collaboration with expert $E_2$.
\lcx{In this case study, $E_2$ evaluated the effectiveness of CNNVis on a set of variants of BaseCNN (with different depths and widths) qualitatively based on his experience.
He also checked the possibility to select a CNN with a suitable architecture under the guide of CNNVis. Though a lot of high-performance models can be referred to on benchmark datasets, it usually takes a long time to transfer the experience to other scenarios (e.g., choose a suitable CNN on a new dataset).
Therefore, $E_2$ emphasized that a \doc{systematic} study on the network architecture and its influence on the performance is necessary to summarize reusable knowledge from existing trials and hopefully transfer it to the development process of other relevant deep models.\looseness=-1
}


\noindent\textbf{\normalsize Overview of BaseCNN.}
We first provided expert $E_2$ with an overview of BaseCNN (Fig.~\ref{fig:basecnn}) to evaluate the quality of CNNVis.

From the overview, he identified that the neurons in the lower layers learned to detect simple patterns such as corners, color patches, and stripes (Fig.~\ref{fig:basecnn}A).
\lcx{A similar observation was reported in previous work~\cite{Krizhevsky2012_NIPS}.}
He identified a neuron for detecting a color patch in a cluster that mainly consists of neurons for detecting stripes with various orientations.
To better compare the neurons that detect color patches, he dragged the neuron to a cluster that mainly consists of neurons for detecting color patches (Fig.~\ref{fig:basecnn}B).
Switching between the top-5 image patches that highly activate a given neuron in lower layers (Fig.~\ref{fig:basecnn}A), he noticed that the retrieved patches did not show much difference in appearance.
Then he turned to higher layers.
After exploring among the top-5 image patches for a given neuron in higher layers (Fig.~\ref{fig:basecnn}C), he noticed that these neurons could learn to detect more abstract features (e.g., an automobile).
He concluded that, ``The ability of detecting more abstract features in the higher layers is a nice property of well-trained deep CNNs
\lcx{and CNNVis indeed shows this pattern well}.''

\lcx{To further evaluate the ability of CNNVis to visualize the finer details of CNNs, $E_2$ selected two similar classes (automobile and truck) and
then examined the activation patterns of the relevant neurons.}
From the learned features in the lower layers, he found some common parts of trucks and automobiles, such as wheels (A1, A2 in Fig.~\ref{fig:depth} (a)).
He indicated that \lcx{these features are not sufficient to distinguish these two classes.}
Thus, he expanded the \lcx{4-th group of} convolutional layers for further examination (Fig.~\ref{fig:depth} (b)).
Expert $E_2$ noticed that the number of ``impure'' neuron clusters gradually decreases \lcx{as he moved to the higher layers}.
Here, an ``impure'' neuron cluster means that the \junzz{image patches} that maximally activate the neurons in the cluster are from different classes.
\sjx{Examining the ``purity'' means that we check the ability of a CNN to distinguish different semantics conveyed by class labels.
In a pure cluster, the image patches that have the same semantics (class label) are gathered together in the activation space generated by the outputs of the layer.
Note that in the lower layers, we prefer ``impure'' clusters because we want the neurons to detect as many different kinds of features as possible.
While in higher layers, we prefer ``pure'' clusters because we want the model to separate higher-level semantics (different classes) by a large margin, so \junzz{that the} image patches from different classes seldom exist in the same cluster.
We illustrate this criterion in Fig.~\ref{fig:pure}.
For example, in the top convolutional layer of BaseCNN, all clusters look \junzz{``pure"}, which indicates that \junzz{the} output activations given by BaseCNN match well with the semantics of different classes.}



\begin{figure}[ht]
  \centering
  \includegraphics[width=\linewidth]{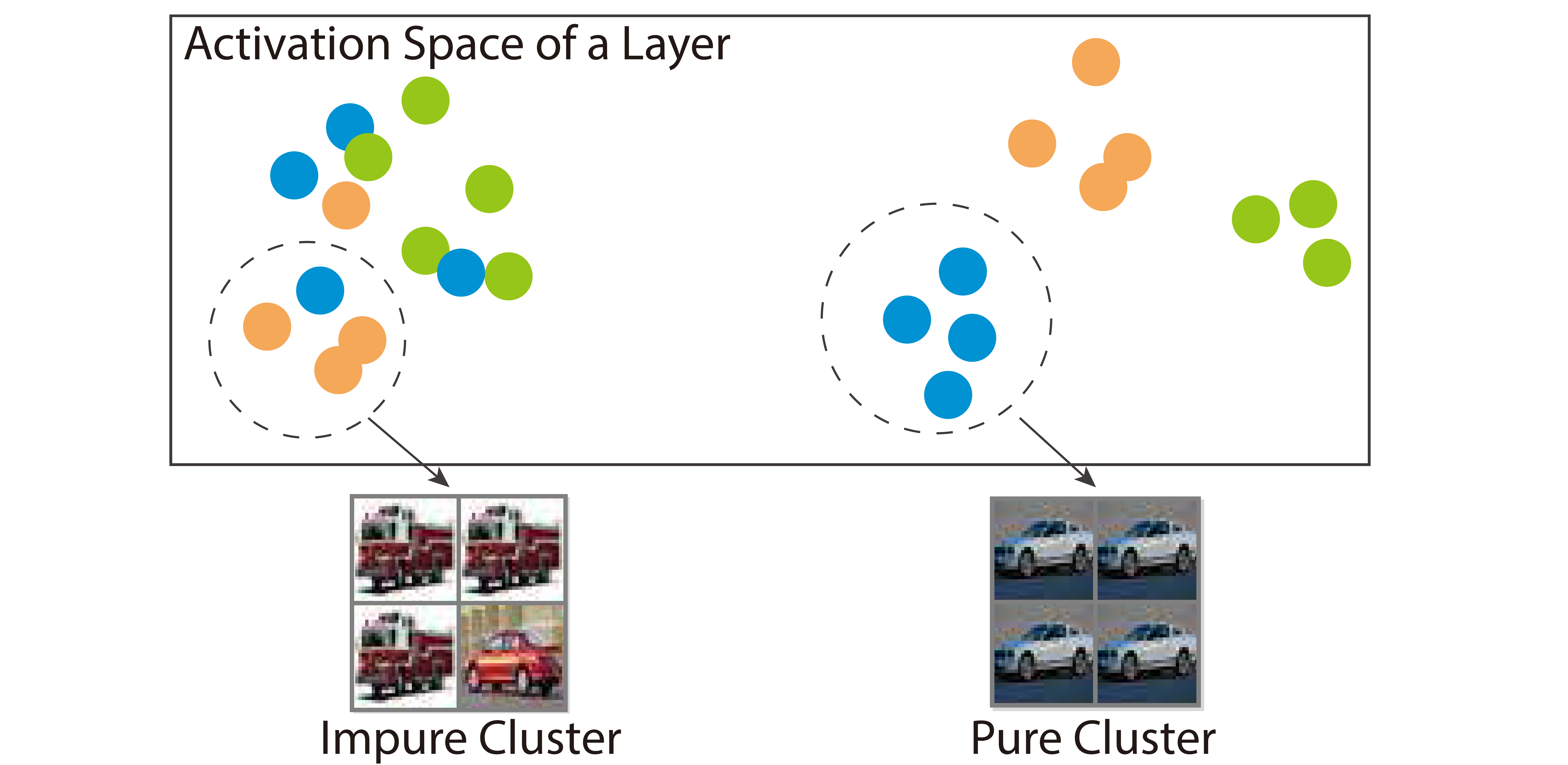}
  \vspace{-5mm}
  \caption{Illustration of an ``impure'' cluster and a ``pure'' cluster.}\looseness=-1
  \label{fig:pure}
  \vspace{-4mm}
\end{figure}

\noindent\textbf{\normalsize Network Depth.}
$E_2$ further investigated how the depth of the network affects the features detected by the neurons. 
He compared BaseCNN with two variant models, including ShallowCNN, which cuts off the 4-th group of convolutional layers, and DeepCNN, which doubles the number of convolutional layers.
The architectures and accuracies are summarized in Table~\ref{table:case1_depth}.
He also selected the truck and automobile \lcx{classes}, and expanded the last group of convolutional layers (Fig.~\ref{fig:case1shallow} (a)).
In ShallowCNN, he identified that there were indeed a lot more ``impure'' clusters in the top convolutional layers compared to those in BaseCNN,
which indicates that a model without a sufficiently large depth is often
incapable of distinguishing the images from similar classes, which
can lead to a \junzz{decrease} of the performance.
In DeepCNN, expert $E_2$ noticed that almost all the weights in the first convolutional layer in the 4-th group were positive (Fig.~\ref{fig:case1shallow} (b)).
The expert commented that since the inputs of that layer were non-negative, the outputs are mostly positive.
The outputs are then fed into ReLU.
As ReLU retains a positive part of the inputs, the ReLU layer, together with its corresponding convolutional layer, can be viewed as a close-to-linear function.
By further expanding the 4-th group of convolutional layers, expert $E_2$ identified several consecutive layers that have a similar pattern (Fig.~\ref{fig:case1deep}).
Because the composition of linear functions is still linear, he concluded that this phenomenon indicates redundancy in the layers.
He also commented that such redundancy may hurt overall performance and make the learning process computationally expensive and statistically ineffective.
These findings are consistent with previous research~\cite{Sun2016_AAAI}.
$E_2$ then concluded that CNNVis could be used to check the abstractness of the features extracted by CNNs.

\begin{table}[ht]
\vspace{-3mm}
    \caption{Performance comparison between CNNs with different depth. ``$\#$ConvLayers'' is the number of convolutional layers and ``$\#$Layers'' is the number of layers that can be visualized.}\label{table:case1_depth}
    \centering \scalebox{0.9}{
    \begin{tabular}{|c|c|c|c|}
    \hline
                                      & Error         & $\#$ConvLayers  & $\#$Layers\\
    \hline
    ShallowCNN        & 11.94\%   &    7       &  30\\
    BaseCNN                      & 11.33\%   &    10  & 40\\
    DeepCNN                      & 14.77\%   &    20 & 70 \\
    \hline
    \end{tabular}
    }
    \vspace{-1mm}
\end{table}

\begin{figure}[ht]
  \centering
  \includegraphics[width=\linewidth]{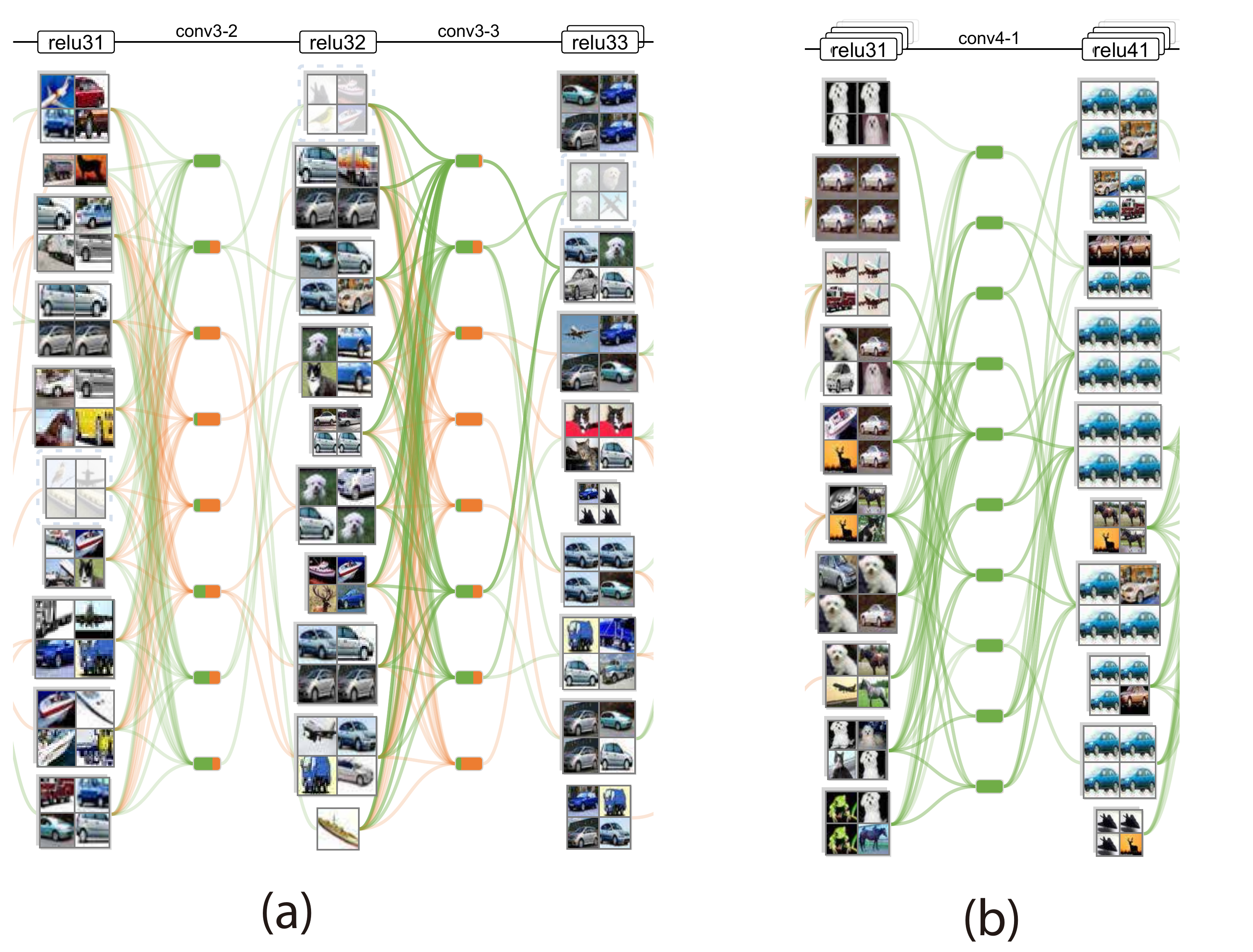}
  \vspace{-5mm}
  \caption{Influence of the model depth: (a) high level features of a shallow CNN; (b) A convolutional layer whose weights are almost positive in DeepCNN.}\looseness=-1
  \label{fig:case1shallow}
  \vspace{-3mm}
\end{figure}

\begin{figure}[ht]
  \centering
  \includegraphics[width=\linewidth]{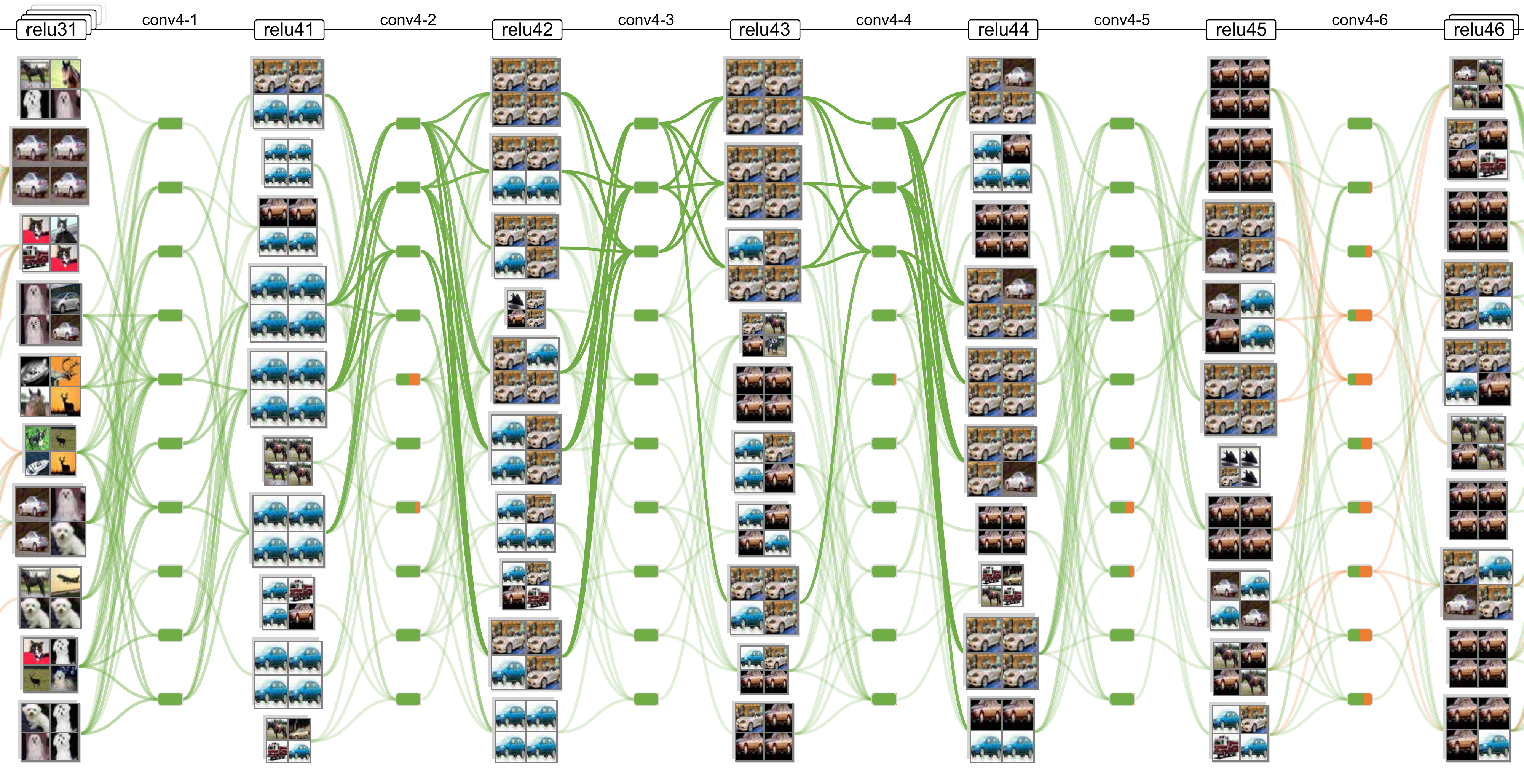}
  \vspace{-5mm}
  \caption{Consecutive convolutional layers whose weights are almost positive in DeepCNN.}\looseness=-1
  \label{fig:case1deep}
  \vspace{-3mm}
\end{figure}




\noindent\textbf{\normalsize Network Width.} Another important factor that influences performance is the width of a CNN.
To have a comprehensive understanding of its influence, $E_2$ evaluated several variants of BaseCNN with different widths, named \junz{by} BaseCNN$\times w$,
where $w$ denotes the ratio of the number of neurons in a layer compared \junz{to} that of BaseCNN.
For example, BaseCNN$\times$4 contains four times the neurons of BaseCNN.
In the case study, $w$ is selected from $\{4, 2, 0.5, 0.25\}$.
The architecture and performance of these variants as well as BaseCNN are listed in Table~\ref{table:case1_width}.

\begin{table}[ht]
\vspace{-3mm}
    \caption{Performance comparison between CNNs with different widths. $\#$params is the number of parameters in the model, which is measured in millions.}\label{table:case1_width}
    \centering \scalebox{0.9}{
    \begin{tabular}{|c|c|c|c|c|c|}
    \hline
                                      & Error         & $\#$params  & Training loss & Testing loss\\
    \hline
    BaseCNN$\times$4        & 12.33\%   &    4.22M        &   0.04       & 0.51\\
    BaseCNN$\times$2        & 11.47\%   &    2.11M        &   0.07       & 0.43\\
    BaseCNN                      & 11.33\%   &    1.05M        &   0.16         & 0.40\\
    BaseCNN$\times$0.5     & 12.61\%   &     0.53M       &    0.34        & 0.40\\
    BaseCNN$\times$0.25   & 17.39\%   &     0.26M        &   0.65       & 0.53\\
    \hline
    \end{tabular}
    }
    \vspace{-1mm}
\end{table}

Compared to BaseCNN, \junz{a} wider network (BaseCNN$\times$4) has a much lower training loss than testing loss.
The expert commented that this phenomenon is known as overfitting in the field of machine learning.
It means that the network \junz{tries} to model every minor variation in the input, which is more likely to be noise.
It often occurs when we have too many parameters relative to the number of training samples.
When a model \junz{overfits}, its performance on the testing set will be much worse than that on the training set.
$E_2$ wanted to examine the influence of overfitting on CNNs.
He visualized BaseCNN$\times$4 with our visual analytics system.

After examining the higher level features, the expert did not found \junz{much} difference compared to BaseCNN.
Then he switched to examine low level features.
He instantly found \junz{that} several neurons learn to detect almost the same features (Fig.~\ref{fig:case1widenarrow} (a)).
The expert inferred that there may be redundant neurons in an \junz{overfitting} CNN.
For further verification, he decided to examine the activations of the neurons in this cluster.
Compared to the activations in lower layers of BaseCNN (Fig.~\ref{fig:case1widenarrow} (b)), he found that many neurons have very similar activations.
This observation verified that there are redundant neurons in the lower layers of a CNN that is too wide.

$E_2$ commented, ``We often use a quantitative criterion (e.g., accuracy) to evaluate the quality of a model.
However, a quantitative criterion itself cannot provide sufficient intuition and clear guidelines.
Even I know a CNN overfits, it is hard to decide which layer to narrow down or remove.
While CNNVis can guide me to locate the candidate layers, which is very useful in my research.''

$E_2$ then compared the performance of BaseCNN with narrower networks (BaseCNN$\times$0.5 and BaseCNN$\times$0.25).
Although the training loss and testing loss of these narrower networks are comparable, which indicate \junz{that these narrow networks generalize well}, their performance \junz{was} worse than BaseCNN (Table~\ref{table:case1_width}).
The expert explained that this phenomenon is known as underfitting.
It happens when the task is complex but we are trying to use a simple model to perform the task.
In image classification, one of the major disadvantage of underfitting is that the model is too simple to distinguish images from similar classes (e.g., automobiles and trucks).
In addition to the decrease in accuracy, he wanted to know the influence that underfitting brought to the model. 

The expert visualized BaseCNN$\times$0.25 for further exploration.
He selected two similar classes, \junzz{automobile} and \junzz{truck}, to examine the patterns of the relevant neurons.
After analyzing low level features, he did not find much difference compared to BaseCNN.
Thus, he switched his attention to high level features.
When examining the features of the last convolutional layer, he found that there were several ``impure'' neuron clusters.
For example, cluster C in Fig.~\ref{fig:case1widenarrow} (c) is represented by three trucks and an automobile (outlier).
He switched to explore the activations in this cluster (Fig.~\ref{fig:case1widenarrow} (c)).
The expert found the outlier has similar activations on the two classes (i.e., \junzz{truck} and \junzz{automobile}),
which means that this neuron can hardly distinguish automobiles from trucks.
As a result, the ability of the model to correctly classify images from similar classes is hindered, which is reflected in the decrease of accuracy.\looseness=-1

Expert $E_2$ commented that, ``\lcx{It is really hard for me to choose the architecture, including the depth and width of the network on a new dataset,
as there are not many high-quality deep models to refer to.
I usually need to try a series of parameters to achieve a satisfactory performance.
CNNVis can \junz{intuitively} show the quality of the model in various ways, such as the purity of clusters,
and help me find the suitable architecture more quickly.}''

\begin{figure}[ht]
  \centering
  \includegraphics[width=\linewidth]{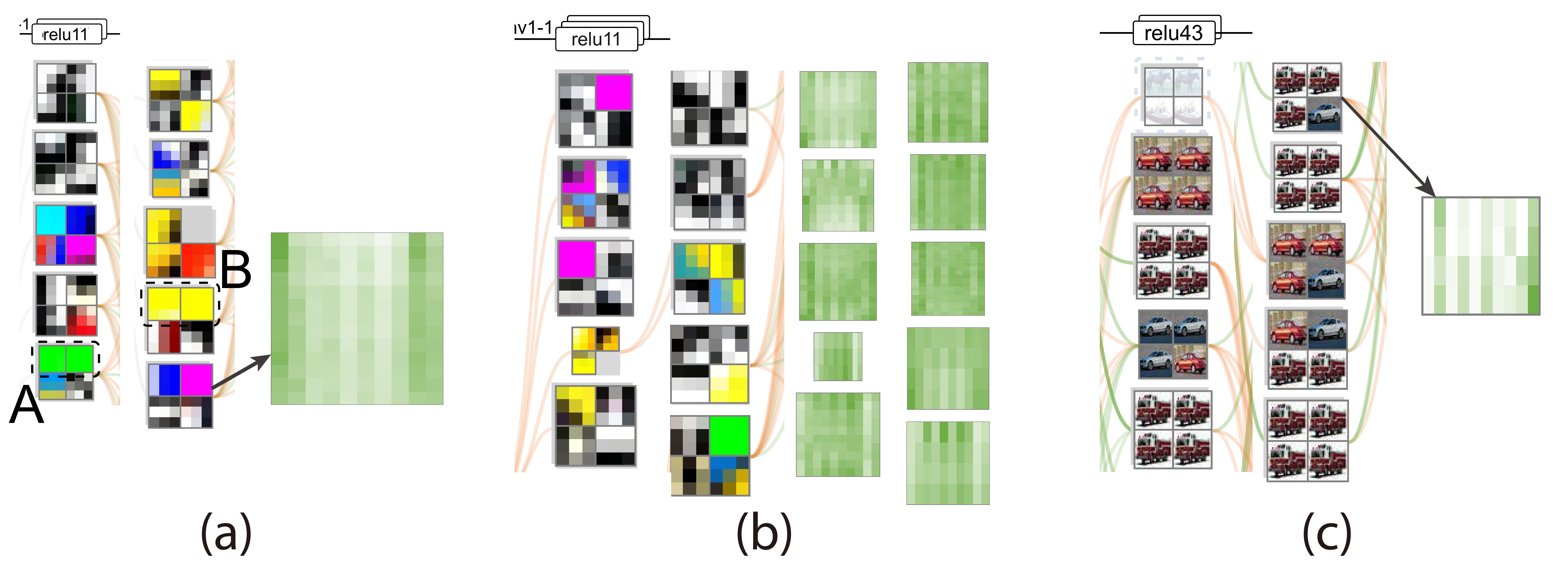}
  \vspace{-7mm}
  \caption{Comparison between models with different widths: (a) low level features of BaseCNN$\times$4; (b) low level features of BaseCNN; (c) high level features of BaseCNN$\times$0.25}\looseness=-1
  \label{fig:case1widenarrow}
  \vspace{-3mm}
\end{figure}

\subsection{Case Study: Training Diagnosis}
\label{sec:case2}

\begin{figure}[ht]
  \centering
  \includegraphics[width=\linewidth]{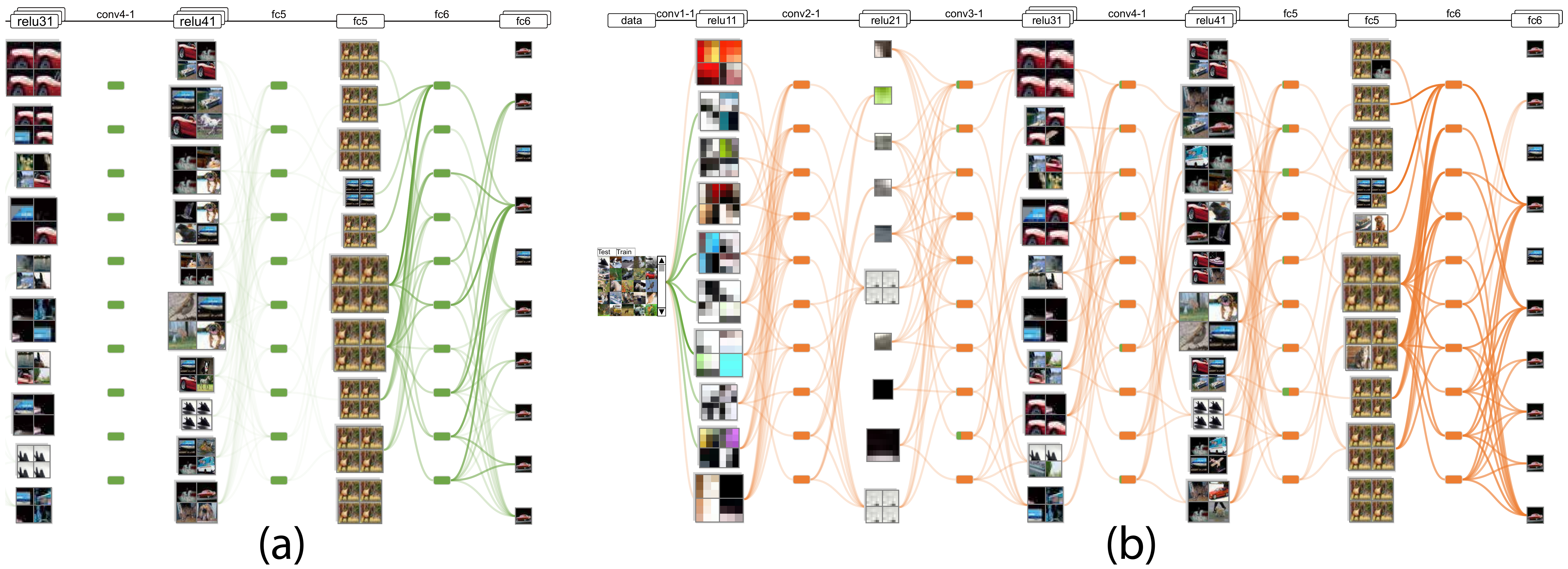}
  \vspace{-7mm}
  \caption{Exploring the connections between neurons: (a) \sjx{edges encoded by} the relative change of weights; (b) edges encoded by the weights.}\looseness=-1
  \label{fig:case2-1}
  \vspace{-3mm}
\end{figure}

\begin{figure}[ht]
  \centering
  \includegraphics[width=\linewidth]{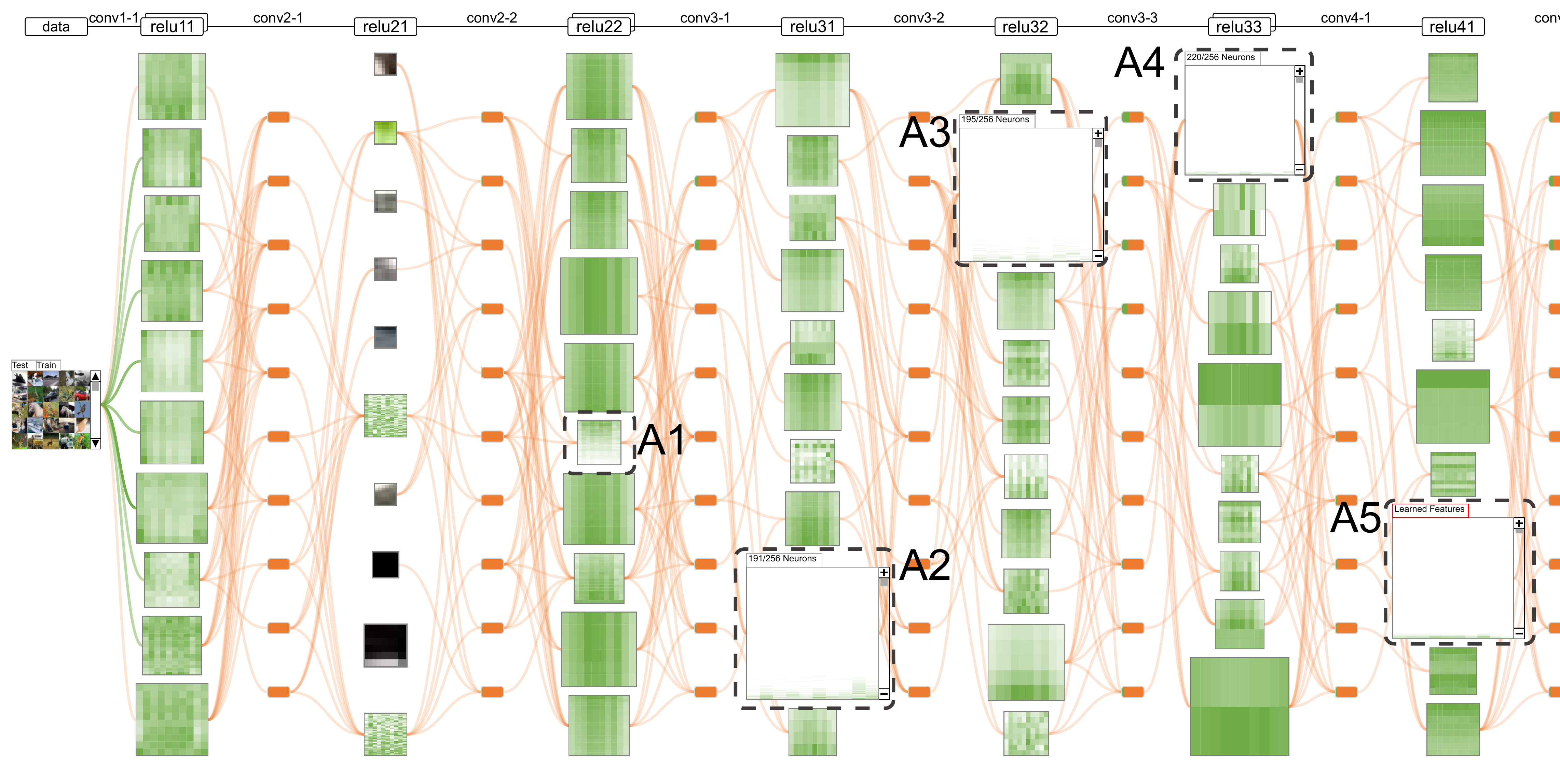}
  \vspace{-7mm}
  \caption{Exploring the neuron clusters.}\looseness=-1
  \label{fig:case2-2}
  \vspace{-2mm}
\end{figure}

This case study demonstrates how CNNVis helps an expert ($E_3$) diagnose a failed training process.
Recently, during the research triggered by~\cite{li2015max}, $E_3$ tried to construct a variant of BaseCNN.
Specifically, he replaced the output activation function with the identity function (i.e., $f(x)=x$) and the loss function with the hinge loss (see the loss function part in Sec.~\ref{sec:background}).
However, the training of this model failed.
The problem was that the training process got stuck when the loss decreased to around $2.0$, where the model was far from achieving a good accuracy.

To help the expert diagnose the failed training process, we provided him with the visualization of a snapshot after the training process got stuck.
As he often uses the relative changes of weights to diagnose a training process in his previous research,
he set the initial color coding of edges as the relative changes of weights.

From the overview, expert $E_3$ observed that the edges were difficult to recognize after the top-2 layers (Fig.~\ref{fig:case2-1}(a)).
This indicated that the relative changes of weights were very small, which caused the training process being stuck.
$E_3$ was curious about what led to such small relative changes in weights,
so he used the color of edges to represent the weights.
He immediately identified that an overwhelming majority of edges were negative (Fig.~\ref{fig:case2-1}(b)). 

He wanted to find what influence the negative weights had on the model. 
As the learned features could not reveal too much information due to the failed training process, expert $E_3$ switched to examine the activation matrix.
He spotted some neuron clusters where all the neurons had zero activations on all classes.
To further study this phenomenon, he sequentially expanded the second, third, and fourth groups of convolutional layers.
He found that the ratio of neurons with zero activations became larger and larger from the lower layers to the higher layers (Fig.~\ref{fig:case2-2}).
\sjx{The activation functions of these neurons are ReLUs.}
He continued to zoom in and further examine the inputs fed into the ReLUs, which he found were always negative.
If the input of a ReLU is less than zero, it generates a zero activation.\looseness=-1

Expert $E_3$ explained that because the input of each convolutional layer is the output of ReLUs in the previous layer, it must be nonnegative.
As the weights of the linear transformation in this layer are mostly negative, the values fed into ReLUs are mostly negative.
Consequently, the outputs of ReLUs are mostly zeros.
\sjx{In the training method that we used (i.e., stochastic gradient descent \cite{bottou-91c}), zero outputs of a neuron mean zero updates to its weights.}





Having learned why the training process got stuck, expert $E_3$ proposed a method to force the network away from that situation.
He added a batch-normalization layer \cite{ioffe2015batch} after each convolutional and fully-connected layer, before the ReLU activation function.
With batch-normalization, the input fed into the ReLUs should no longer be mostly negative.
This means that the model could still be trained even most weights were negative.

The improved model achieved an average error of 9.43\% on the CIFAR-10 dataset, with which expert $E_3$ was very satisfied.
He further commented, ``I have investigated this problem for a long time and inserted all kinds of code fragments to print the debugging information during training.
However, after many unsuccessful attempts and a great deal of effort spent reading the debugging information, I eventually gave up.
It is awesome to have a toolkit like CNNVis, which intuitively \junzz{illustrates} the training statistics and allows me to explore the training process from multiple perspectives.''

\section{Conclusion}\label{sec:conclustion}

In this paper, we have presented a novel visual analytics system to help machine learning experts better understand, diagnose, and refine CNNs.
Powered by a hybrid visualization consisting of rectangle packing, matrix ordering, and biclustering-based edge bundling, the system allows experts to explore and understand a deep CNN from different perspectives.
In addition, it enables experts to diagnose and refine the CNN architecture to further improve the performance.
Three case studies were conducted to demonstrate the effectiveness and usefulness of the system for comprehensive analysis of CNNs.

There are several directions for future work to further improve our system.
Currently, CNNVis focuses on analyzing a snapshot of the CNN model in the training process, which is useful for conducting the offline analysis.
All the experts expressed the need to integrate CNNVis with the online training process and continuously get an update of the training status.
A key issue is the difficulty of selecting representative snapshots and comparing them effectively.

Another interesting venue for future work is to apply CNNVis to other types of deep models that cannot be formulated as a DAG, such as recurrent neural network (RNN).
The major bottleneck is to design an effective visualization to facilitate experts in understanding the data flow through different types of deep models.
For example, in addition to the conventional multi-layer neural network, RNN has a feedback loop from an output to an input.
Better understanding the working principle of the feedback loop help experts design more effective models.

}

\small
\bibliographystyle{abbrv}
\bibliography{reference}



\end{document}